\documentclass{article}
\usepackage{arxiv}
\usepackage{bm}
\usepackage{amsfonts,amssymb}
\usepackage{amsmath}
\usepackage{graphicx}
\usepackage{subcaption}
\usepackage[detect-all]{siunitx}
\usepackage{balance}
\usepackage{float}
\usepackage{amssymb}
\usepackage{booktabs}
\usepackage{caption,blindtext}
\usepackage{multirow}
\usepackage{url}
\usepackage{color}
\newtheorem{myTheo}{Theorem}

\newtheorem{myLemma}{Lemma}
\title{The Expressivity and Training of Deep Neural Networks: toward the Edge of Chaos?}
\author{
  Gege~Zhang \\
 Department of Automation\\
 Shanghai Jiao Tong University \\
Shanghai 200240, PR China\\
  \texttt{ggzhang@sjtu.edu.cn} \\
\And
   Gangwei~Li\\
NVIDIA Corporation\\
Building 2 No. 26 Qiuyue Road\\
Shanghai PRC, Shanghai 201210, PR China\\
\texttt{vili@nvidia.com} \\
\And
Weining~Shen\\
Information and Computer Sciences\\
 University of California\\
 Irvine, CA 92697, USA\\
 \texttt{weinings@uci.edu} \\
\And
Weidong~Zhang\\
 Department of Automation \\
 Shanghai Jiao Tong University \\
   Shanghai 200240, PR China.\\
  \texttt{wdzhang@sjtu.edu.cn} \\
  }
\begin{document}
\maketitle
\begin{abstract}
Expressivity is one of the most significant issues in assessing neural networks.
In this paper, we provide a quantitative analysis of the expressivity for the deep neural network (DNN) from its dynamic model,
where the Hilbert space is employed to analyze the convergence and criticality.
We study the feature mapping of several widely used activation functions obtained by Hermite polynomials, and find sharp declines or even saddle points in the feature space, which stagnate the information transfer in DNNs. We then present a new activation function design based on the Hermite polynomials for better utilization of spatial representation.
Moreover, we analyze the information transfer of DNNs, emphasizing the convergence problem caused by the mismatch between input and topological structure.
We also study the effects of input perturbations and regularization operators on critical expressivity. Our theoretical analysis reveals that DNNs use spatial domains for information representation and evolve to the edge of chaos as depth increases.
In actual training, whether a particular network can ultimately arrive the edge of chaos depends on its ability to overcome convergence and pass information to the required network depth.
Finally, we demonstrate the empirical performance of the proposed hypothesis via multivariate time series prediction and image classification examples.
%For multivariate time series prediction, the results show that the optimized deep echo state network provides higher predictive performance, especially for long-term prediction.
%For image classification, the convergence increases with the increase of the batch size, but the hidden state is gradually disappearing, indicating that information transfer becomes weak.
\end{abstract}

\section{Introduction}
Deep neural networks (DNNs) have achieved outstanding performance in many fields, from the automatic translation to speech and image recognition \cite{lecun2015deep, silver2016mastering}.
It is now understood that a key ingredient of its success is the so-called {\it high expressivity} property. For example,
Bianchini, Scarselli and Zhang et al. \cite{Bianchini2014On, ZhangWCLMSB16} showed by numerical analysis that DNN had an exponential expressivity as depth increased.
Similar results were obtained using a topological measure in \cite{poole2016exponential}.
Efforts have also been made to explore the black box structure of DNNs from a theoretical perspective.
This starts with the investigation of the dynamics (e.g., depth and width) of DNNs; and it was found that the dynamics of DNNs were very different from those of shallow networks  \cite{bartlett1999almost}.
Subsequently, mathematical analysis, e.g., wavelet and function approximation frameworks, were applied towards a better understanding of  DNNs, although most of the works were limited to two-layer networks only \cite{Mallat2016, ZhangBHRV17}.
Further, due to the non-linear functional mappings of the input-to-output activation functions, the convergence and expressivity with respect to activation functions have been investigated \cite{AgostinelliHSB14, RaghuPKGS17, LiY17, tian2017analytical, YunSJ19a}.
Although these studies provide significant advances in understanding DNNs, their excellent representation performance is still largely not well understood.
Moreover, as noted by many previous works, training DNNs remains very challenging, where the major difficulties include the existence of local minimums, low curvature regions, and exponential growth or fading of backpropagation gradients  \cite{Shen18}.

To solve the aforementioned challenges in interpreting and implementing DNNs, the current development of complexity theory prompts people to understand DNNs from a general theoretical perspective.
It has been pointed out that DNNs could be thought of as a type of complex network that possesses an exponential expressivity in their depth, where the
information capacity reaches maximum in an order-disorder phase transition \cite{poole2016exponential, RaghuPKGS17, Stanley1973Introduction}.
All features of interest can be represented by a medium-sized network, similar to today's DNNs, etc.
This motivates the relevant literature on investigating DNNs from a complexity theory perspective.
For example, Hanna et al. \cite{Hanna2015The} established a natural connection between the autoencoder and the energy function,
 and then provided a classifier based on class-specific autoencoder.
Morningstar et al. \cite{MorningstarM17} showed that shallow networks were more effective than DNNs in representing probability distribution.
Del et al. \cite{Del2017Criticality} investigated the critical characteristics of self-organizing recurrent neural network (RNN),
and showed that a neural plasticity mechanism was necessary to reach a critical state, but not to maintain it.
Moreover, field theories such as Mean-field, Symmetry Breaking, and Renormalization Group have been postulated to explain the criticality of DNNs \cite{poole2016exponential, yang2017mean, hartnett2018replica, Kochjanusz2017Mutual, ChenPS18}.
Some network structures such as small-world networks and scale-free networks have been utilized for the topology design of DNNs; and the results indicate that they can achieve faster convergence than that of random networks \cite{wang2019short, monteiro2016model}.
Although the previous work provides some profound insights into DNNs, a quantitative analysis of the rich dynamics in DNNs remains elusive.

The goal of this paper is to (i) examine the expressivity and training of DNN through quantifying its vanilla version, as well as analyze its convergence and criticality based on its dynamics; and (ii) study the factors influencing the convergence of DNNs such as activation functions, inputs, and model training.
These results will provide more insights on how to improve the empirical performance of DNNs. We summarize our major contributions as follows:
\begin{itemize}
 \item In theory, we show that DNN uses spatial domains for information representation and approaches the edge of chaos as depth increases.
In actual training, whether a network can ultimately arrive at the edge of chaos or not depends on its ability to overcome convergence and pass information to the desired network depth.
 \item By analyzing the feature mapping of commonly used activation functions under Hermite polynomials, we find that there are significant drops or even saddle points in the feature space that slow down the message passing in DNNs.
 We propose an activation function design based on the Hermite polynomials to accelerate convergence.
 \item We also analyze the information transfer in DNNs with an emphasis on model matching between inputs and topological structures. The results suggest that the appropriate network size is crucial to reach the chaotic edge.
 \end{itemize}

The rest of the paper is organized as follows. Sec. \ref{DNNBasics} analyzes the dynamic model of a vanilla DNN.
Sec. \ref{HilbertFeatureMapping} studies the feature mapping of several commonly used activation functions under Hermite polynomials and proposes a new activation function to use.
Sec. \ref{networksModelsTraining} examines information transfer in model training, focusing on the model matching between inputs and topological structures.
Sec. \ref{InputPerturbationandRegularization} studies the impact of input perturbation and regularization on critical representation.
Sec. \ref{modelValidity} verifies the provided perspective by multivariate time series prediction and image classification.
We discuss some relevant work in Sec. \ref{relatedWorks} and conclude with a few remarks in Sec. \ref{Conclusions}.

\section{Dynamics of a vanilla deep neural network}\label{DNNBasics}
In this section, we study the convergence and criticality properties of DNN based on its dynamic model. First consider a vanilla DNN with an $L$-layer weight matrix $\bm{W}_1,\dots,\bm{W}_L$ and $(L+1)$-layer
neural activity vectors $\bm{x}_0, \dots,\bm{x}_L$, assuming $N_l$ neurons in the layer $l$, so that $\bm{x}_l \in {\mathbb{R}}$ and $\bm{W}_l$ is a $N_l\times N_{l-1}$ weight matrix.
The feedforward dynamics from the input $\bm{x}_0$ is given by
\begin{equation}\label{eq:dynamics}
\bm{x}_l =\sigma(\bm{W}_l \bm{x}_{l-1} + \bm{b}_l), \, \text{ for } l = 1,\dots, L,
\end{equation}
where $\bm{b}_l$ is a bias vector, and $\sigma$ is an activation function that transforms inputs to nonlinear outputs. Without loss of generality, we assume that the weight matries $ \bm {W}_l $ all follow a normal distribution and denote $\bm{W}_l$ as $\bm{W}$ for simplicity.

Criticality plays a significant role in the theory of phase transitions.
 It appears in a variety of complex networks, e.g., water freezing and iron magnetizing \cite{Stanley1973Introduction}.
As a network approaches a critical point, it exhibits some interesting phenomena, such as fantastic expressivity and self-organization, etc.
DNNs maximize entropy between inputs and outputs through information transfer between neurons, which are analogous to complex networks.
Below we analyze the criticality of a vanilla DNN from dynamics by utilizing its Lyapunov function (the eigenvalues of the system's Jacobian matrix).
According to the chain rule, the Jacobian matrix of Eq. (\ref{eq:dynamics}) can be expressed as
\begin{equation}\label{eq:JacobianMatrix}
{\bm{J}}({\bm{x}})=\sigma^{\prime}(\bm{W} \bm{x} + \bm{b})\bm{W},
\end{equation}
where $\sigma^{\prime}$ is the derivative of activation function $\sigma$. Then the network status can be calculated as the eigenvalues of the Jacobian matrix at time $n$:
\begin{equation}
\lambda=\max\limits_{l=1,\dots,L} \frac{1}{N}\sum_{n=1}^N \textrm{log}(|\sigma_l(n)|),
\end{equation}
where $\sigma_l(n)$ denotes the eigenvalues of ${\bm{J}}(\bm{x})$ at time $n$.
There may be many equilibrium points for general DNNs.
To determine whether an equilibrium point is stable, we can check the local approximation at the equilibrium point. For instance, by linearing \eqref{eq:dynamics} at the equilibrium point $\bm{x}_0$, we obtain
\begin{equation} \label{eq:LinearizeAproximate}
{\bm{x}}(n)={{\bm{J}}}({\bm{x}_0})\left(\bm{W}{\bm{x}}\left( {n - 1} \right)\right),
\end{equation}
which is a homogeneous differential equation whose solution can be determined by the roots of its characteristic polynomial. When all roots have negative real parts, the system is considered stable;
it is assumed chaotic when any root posses a positive real part;
and it is on the edge of chaos when any root owns zero real part.
Fortunately, the roots (eigenvalues) of commonly used activation functions are in general constrained within $[0, 1)$, thus avoiding the tendency toward a chaotic state. Moreover, early advances in designing activation function mainly focus on the existence
of global asymptotic/exponential stability of critical points
(along with the lines, concerning the existence, uniqueness, completeness, sparsity, convergence, and accuracy of such a function \cite{zhang2014comprehensive}).
However, it is in general difficult to determine the eigenvalues for an arbitrary dynamic equation.
Therefore, to date, there is still a lack of clear guidelines for designing activation functions.
Therefore, an adaptive activation function called Swish activation has been developed recently and shown to have significantly outperformed the ReLU activation \cite{elfwing2018sigmoid}.

\section{Feature mapping using Hilbert space} \label{HilbertFeatureMapping}
Suitable and universal activation function design can greatly improve the performance of DNNs.
Hilbert space, as a natural infinite-dimensional generalization of Euclidean space, and as such, enjoys the essential features of completeness and orthogonality.
In this section, we present the feature mapping using Hilbert space and then give an appropriate activation function design based on the analysis.
The proofs of all the theorems can be found in Chapters 2 and 5 of \cite{kreyszig1978introductory}.

Consider an inner product space with a norm defined by $\lVert \bm{x}\rVert= \langle x, x\rangle^{1/2}$.
If space $\bm{H}$ is complete with respect to this norm, it is called a Hilbert space. A complete normed vector space is called a Banach space, hence a Hilbert space is a closed subset of a Banach space. The following Contraction-Mapping Theorem, also known as the Banach's Fixed Point Theorem, is a useful tool for guaranteeing the existence and uniqueness of solutions for a differential equation.

\begin{myTheo}\label{FixedPointTheorem} (\textbf{The Contraction-Mapping Theorem})
Given a function $\mathop{T}: X \rightarrow X$ for any set $X$, if an $x \in X$ makes $\mathop{T}(x) =x$, then $x$ is called a fixed point.
If the fixed point is unique,then the only solution can be obtained by the limit of $\bm{x}(n)$ defined by $x(n) = Tx(n-1), n= 1, 2, \dots$, expressed as:
\begin{equation}
x = \lim_{n\rightarrow \infty} T^n x_0,
\end{equation}
where $x_0$ is an arbitrary initial element.
 \end{myTheo}

Theorem \ref{FixedPointTheorem} also provides a useful way to find the solutions to the differential equation through an iterative process. The following is a natural extension of the theorem.
 \begin{myLemma}\label{EigenHilbertSpace}
 If $T$ is a self-adjoint operator, then there exists $B\geq A>0$ satisfying:
 \begin{equation}\label{eq:self-adjoint}
 \forall f \in \bm{H}, \; A\lVert f\rVert^2 \leq \langle \mathop{T}f, f\rangle \leq B\lVert f\rVert^2.
 \end{equation}
If $T$ is invertible, then we have:
 \begin{equation}\label{eq:self-adjoint1}
 \forall f \in \bm{H}, \;\frac{1}{B}\lVert f\rVert \leq \langle {\mathop{T}}^{-1}f, f\rangle \leq \frac{1}{A}\lVert f\rVert^2.
 \end{equation}
 \end{myLemma}

Inequality (\ref{eq:self-adjoint}) shows that the eigenvalues of $\mathop{T}$ are between $A $ and $B$.
 In a finite dimension, it is diagonal on an orthogonal basis since $\mathop{T}$ is self-adjoint.
 It is therefore invertible with eigenvalues between $B^{-1}$ and $A^{-1}$, which proves inequality (\ref{eq:self-adjoint1}).

From the Theorem \ref{FixedPointTheorem} and the Lemma \ref{EigenHilbertSpace}, we know that Hilbert space can convert the dynamics of DNN into a linear model, and the unique solution can be found by increasing depth.
Specifically, the eigenvalues of the current DNNs are within $[-1, 1)$, which make them converge to $0$ by iteration, so that $0$ is a critical point of the system.
From the dynamics in Sec. \ref{DNNBasics}, if any of the roots of  \eqref{eq:LinearizeAproximate} is $0$, then the system is on the edge of chaos.
Therefore, DNN approaches the edge of chaos at an exponential speed as depth increases.
According to the definition of information entropy, that is a property that identifies system's tendency to move toward a completely random or disordered state \cite{latora2000rate}.
 A system of high entropy is more disordered or chaotic than that of low entropy,
therefore, information entropy can be used to the expressivity of DNNs.
From this viewpoint, we also see that the so-called fantastic expressivity in complexity theory will show up when investigating the high-dimensional presentation of DNN.

\begin{figure}
        \centering
        \begin{subfigure}[b]{0.33\textwidth}
            \centering
            \includegraphics[width=\textwidth]{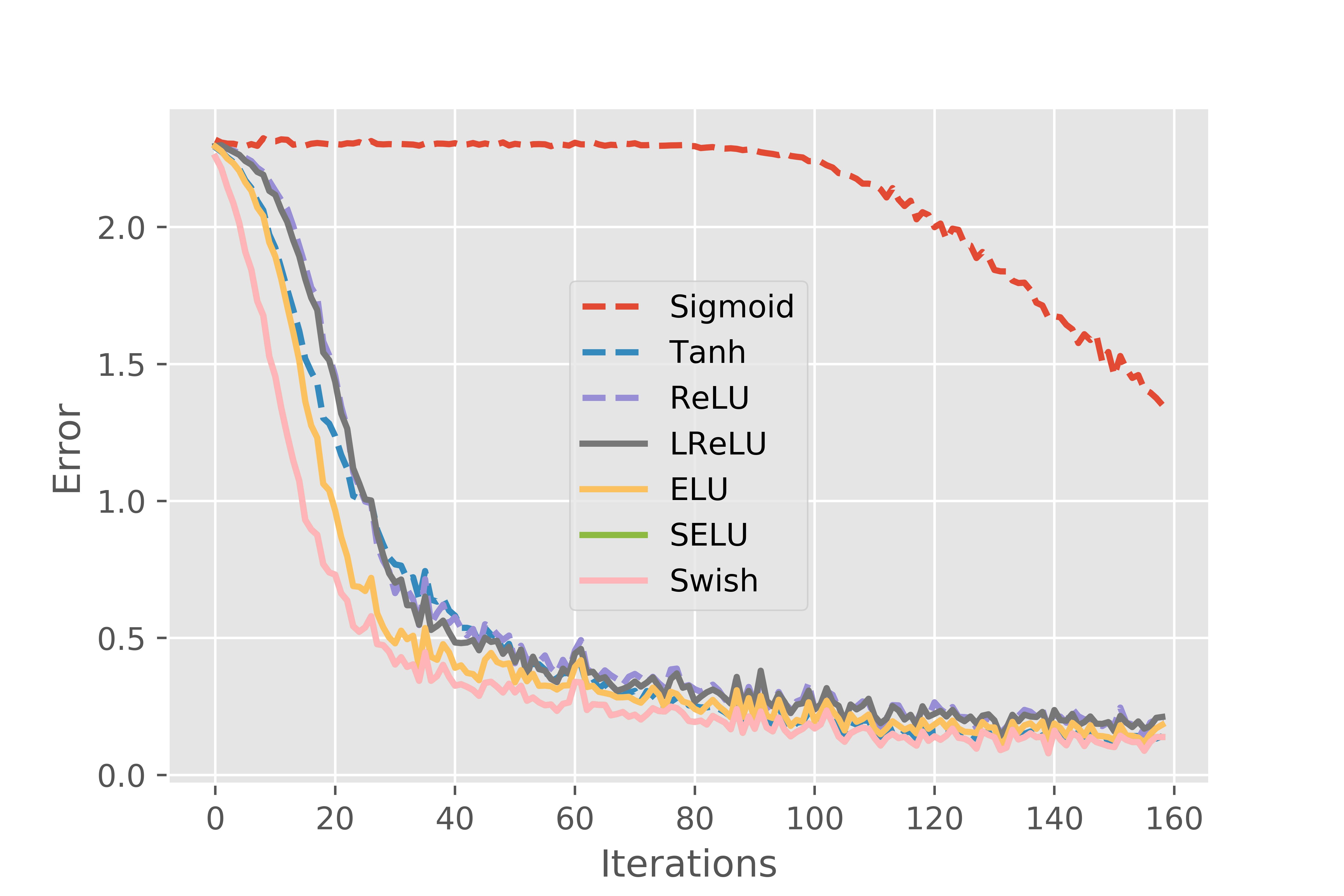}
            \caption[]
            {{\small MNIST data sets}}
        \end{subfigure}
        \begin{subfigure}[b]{0.33\textwidth}
            \centering
            \includegraphics[width=\textwidth]{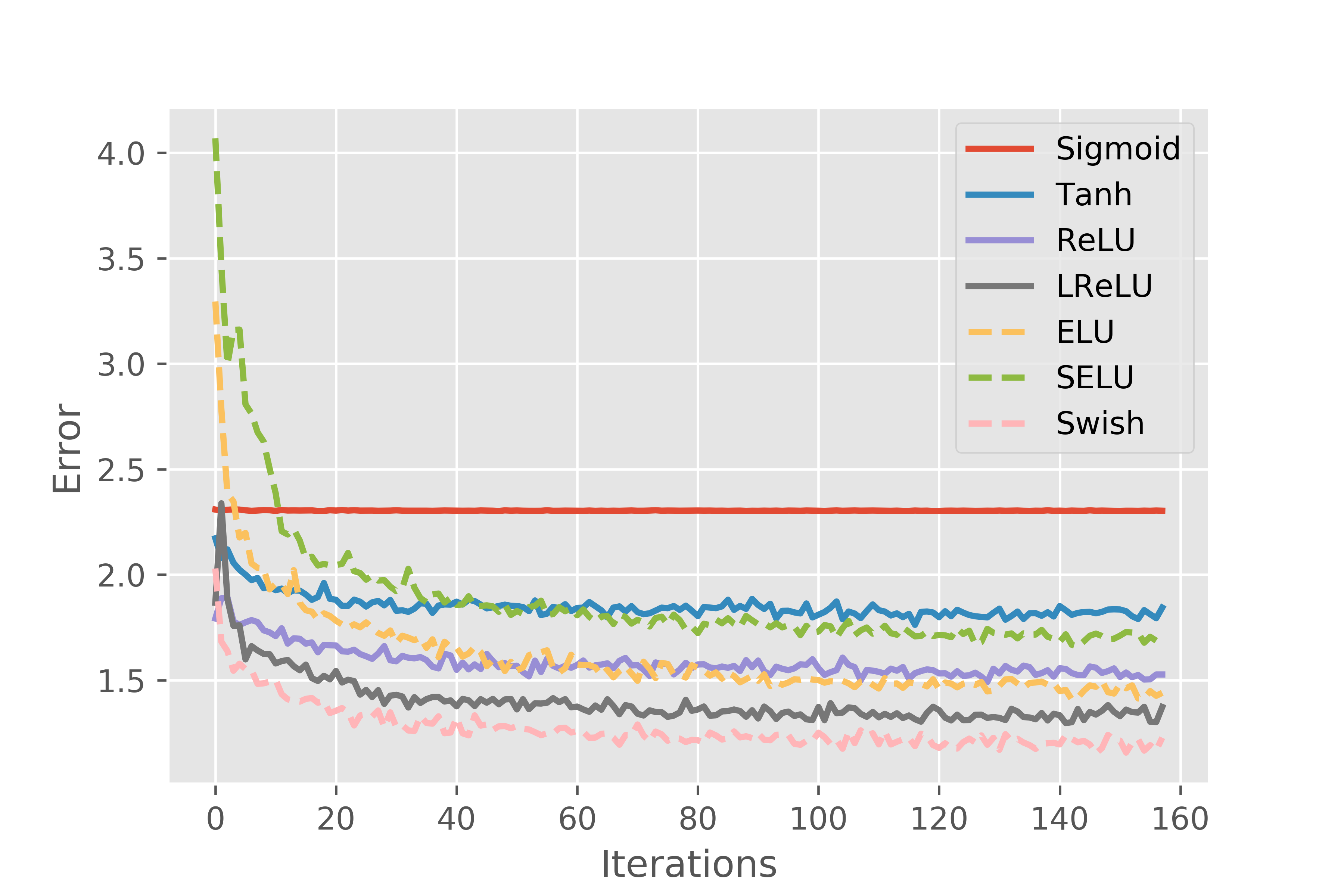}
            \caption[]
            {{\small CIFAR10 data sets}}
        \end{subfigure}
        \caption[]
        {\small \textbf{Comparison of the convergence for different activation functions}.
 The experiment adopted the default settings in the convolutional neural network, including convolution, nonlinearity, and pooling.
The stochastic gradient descent is used as the optimization method. The same experimental setup is used for Fig. 2 and 3.}
        \label{fig:ActivationCompare}
    \end{figure}

From the feature mapping perspective, Hilbert space can be employed to understand the behavior of existing activation functions.
Due to the space limit, we postpone a detailed analysis of the activation functions under Hermite polynomials to \ref{ActivationinHermitePolynomials}.
In Fig. \ref{fig:ActivationCompare}, we provide a numerical comparison of their convergence based on the MNIST and CIFAR10 data sets.
It can be seen that the Sigmoid activation attenuates coefficients with higher amplitude and saturates near the output zero, making it unsuitable for DNNs.
The Tanh activation function saturates at the output $-1$ and $+1$ and has a well-defined gradient at output $0$,
which suggests that the vanishing eigenvalues around output $0$ can be avoided by employing the Tanh activation function.
The ReLU activation achieves fast convergence, and requires cheap computation in implementation.
However, there are many zero elements in the ReLU activation, which correspond to the saddle points in training neural networks. The Swish activation avoids this issue, and it achieves the best results in terms of the fast convergence to the global optima.
The Swish activation largely replaces the Sigmoid activation and the soft Tanh activation because of its easiness in training DNNs.

\begin{figure}
        \centering
        \begin{subfigure}[b]{0.33\textwidth}
            \centering
            \includegraphics[width=\textwidth]{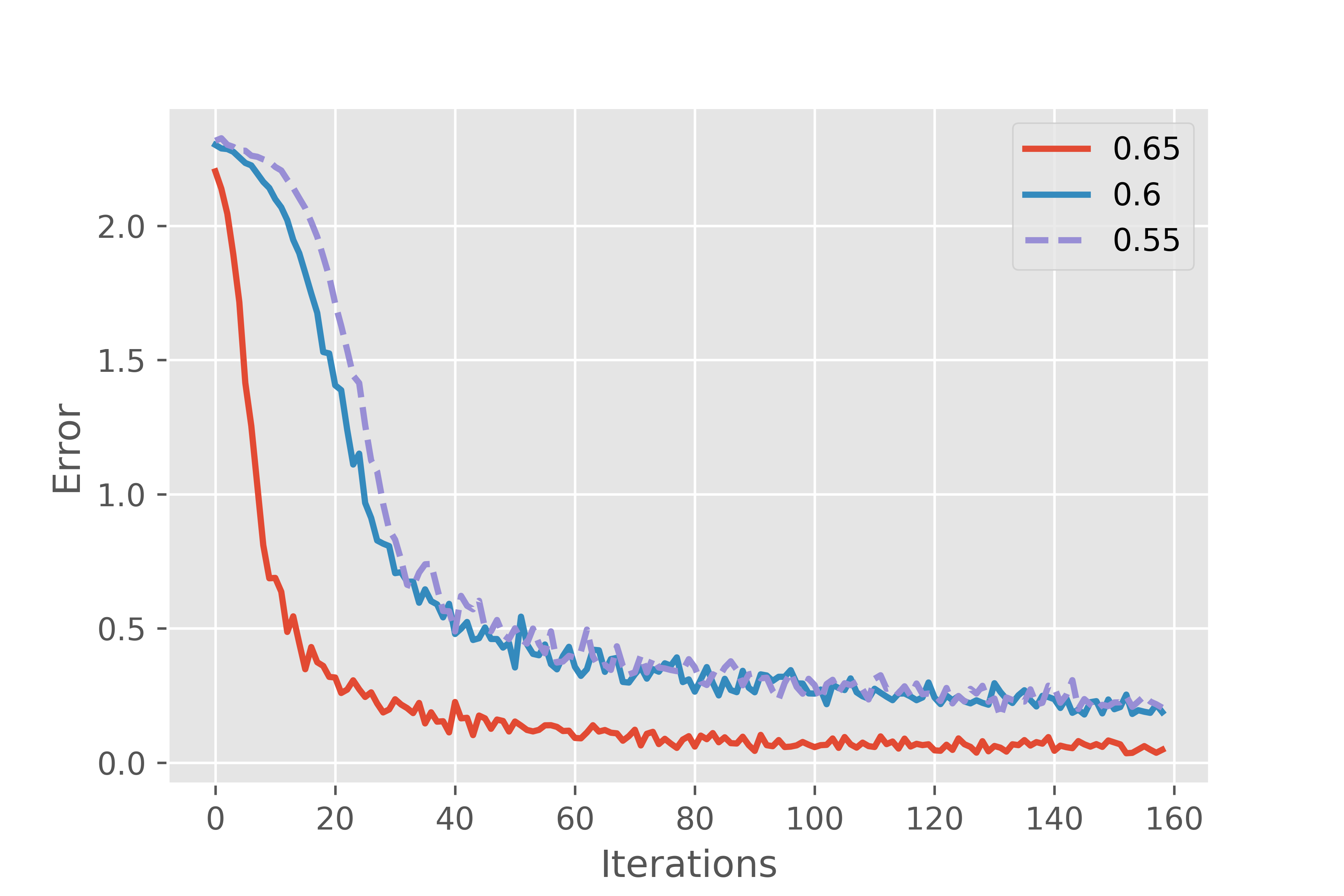}
            \caption[]
            {{\small MNIST data sets}}
        \end{subfigure}
        \begin{subfigure}[b]{0.33\textwidth}
            \centering
            \includegraphics[width=\textwidth]{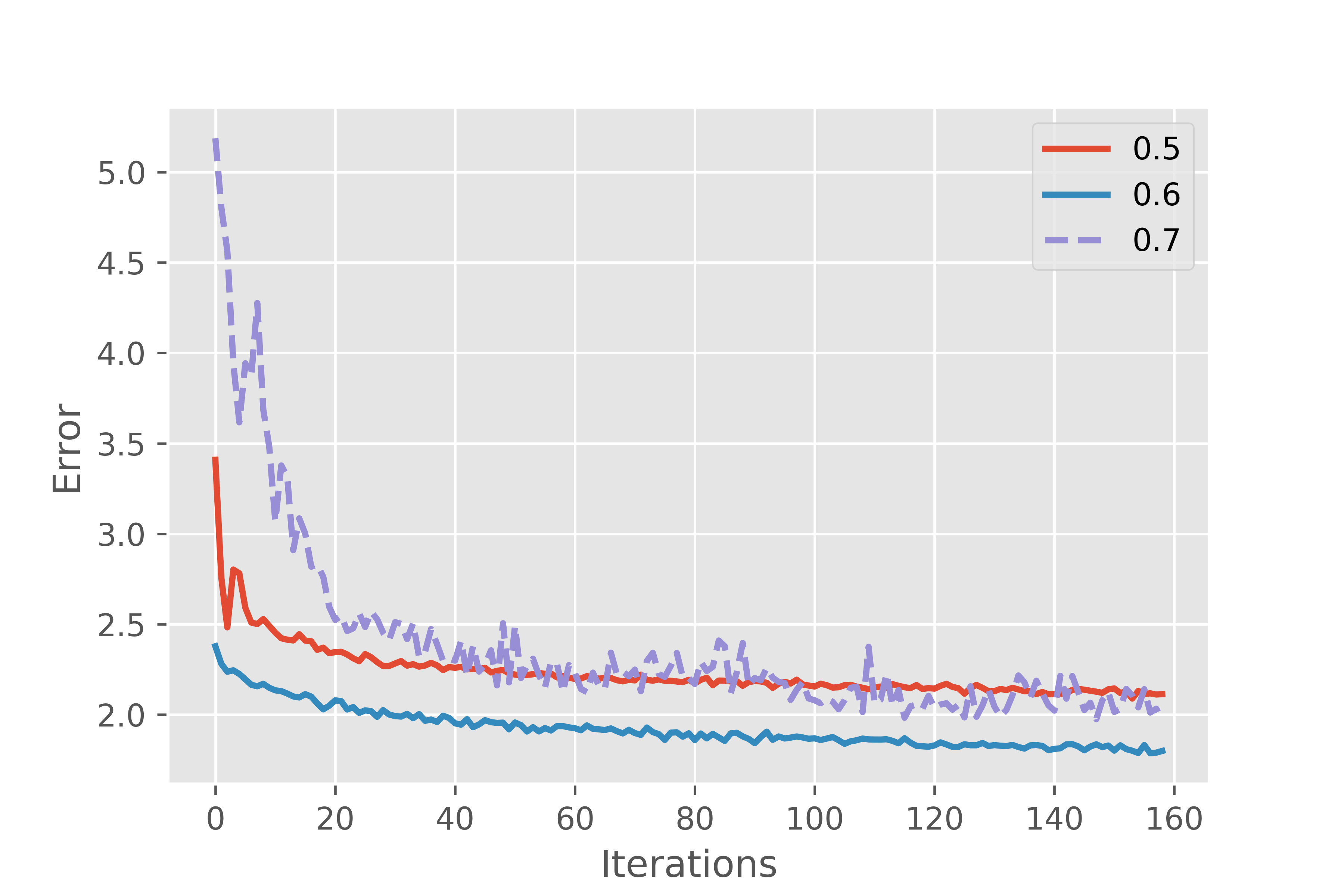}
            \caption[]
            {{\small CIFAR10 data sets  (10 Monte Carlo experiments)}}
        \end{subfigure}
        \caption[]
        {\small \textbf{The maximum Hermite coefficients}.}
        \label{fig:ActivationCompareStart}
    \end{figure}

\begin{figure}
        \centering
        \begin{subfigure}[b]{0.33\textwidth}
            \centering
            \includegraphics[width=\textwidth]{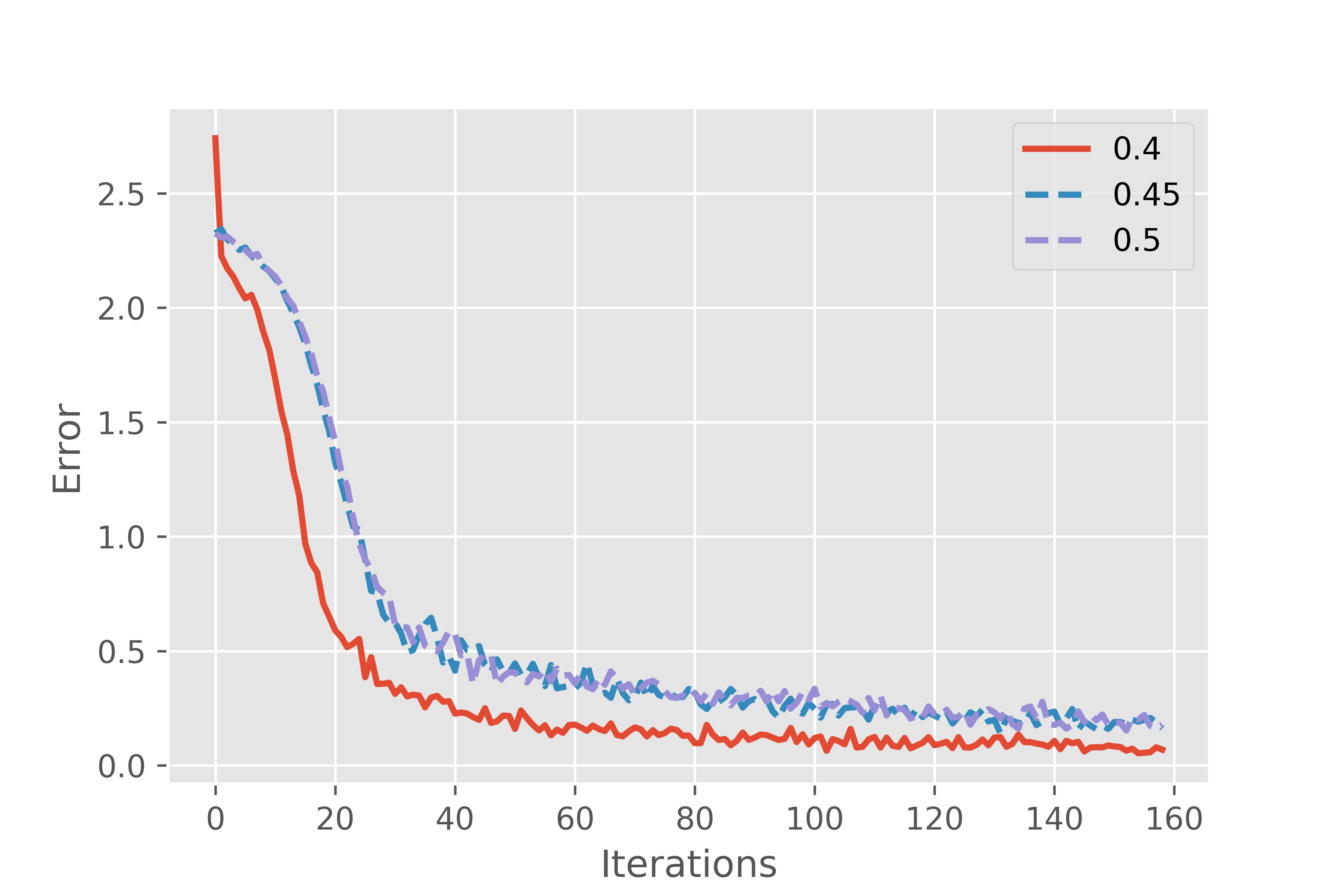}
            \caption[]
            {{\small MNIST data sets}}
        \end{subfigure}
        \begin{subfigure}[b]{0.33\textwidth}
            \centering
            \includegraphics[width=\textwidth]{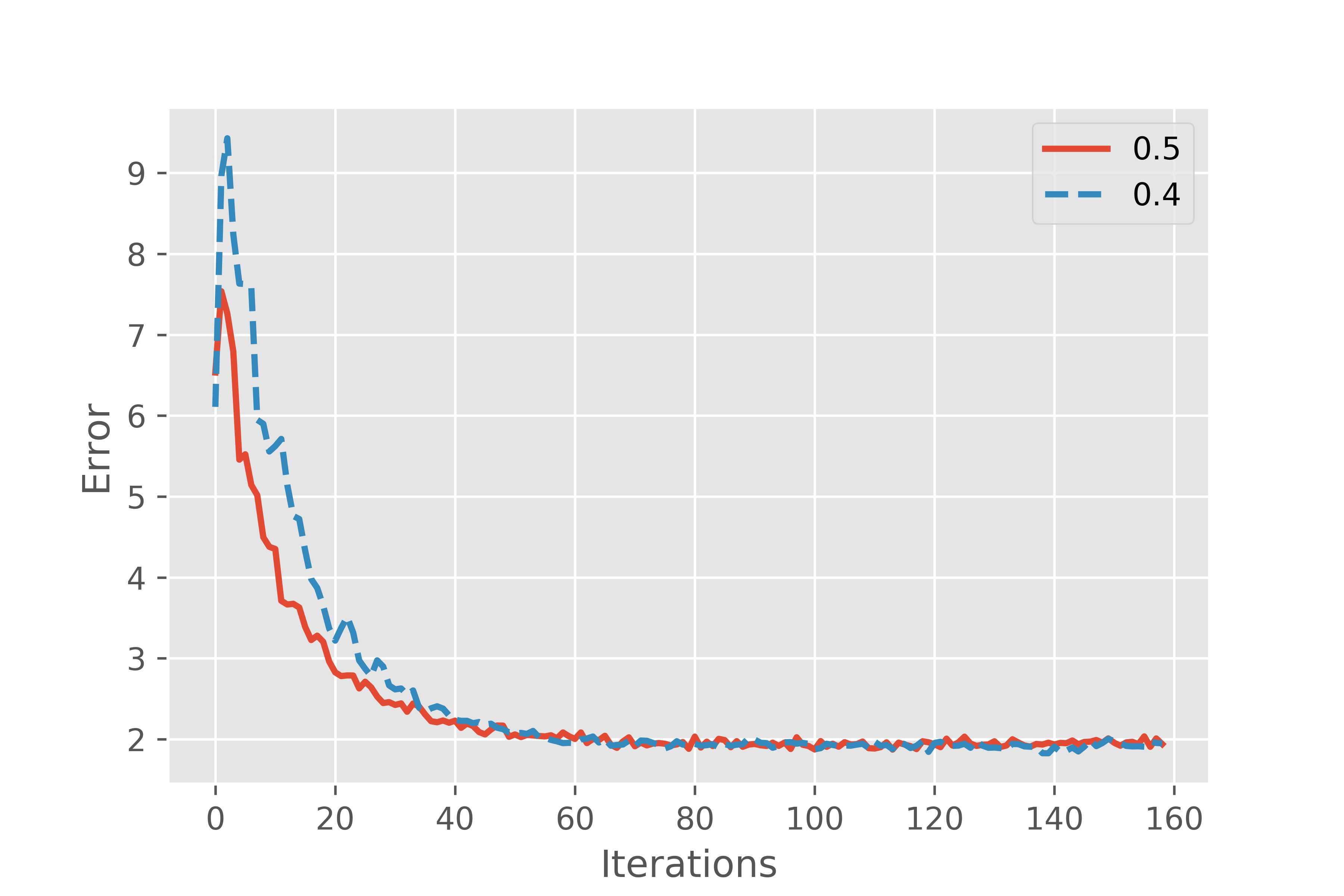}
            \caption[]
            {{\small CIFAR10 data sets (10 Monte Carlo experiments)}}
        \end{subfigure}
        \caption[]
        {\small \textbf{The minimum Hermite coefficients}.}
        \label{fig:ActivationCompareEnd}
    \end{figure}

\begin{figure}
        \centering
        \begin{subfigure}[b]{0.33\textwidth}
            \centering
            \includegraphics[width=\textwidth]{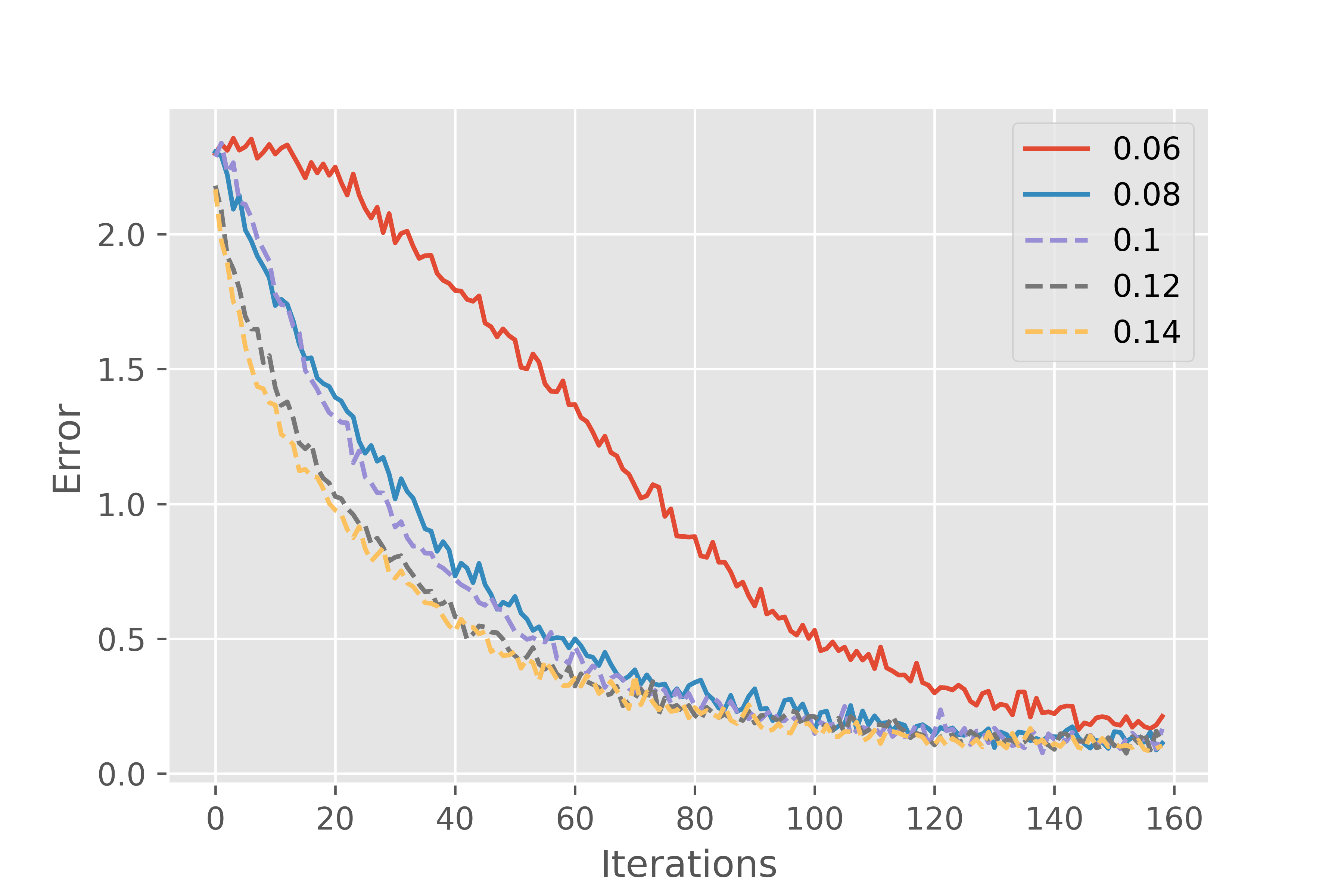}
            \caption[]
            {{\small MNIST data sets}}
        \end{subfigure}
        \begin{subfigure}[b]{0.33\textwidth}
            \centering
            \includegraphics[width=\textwidth]{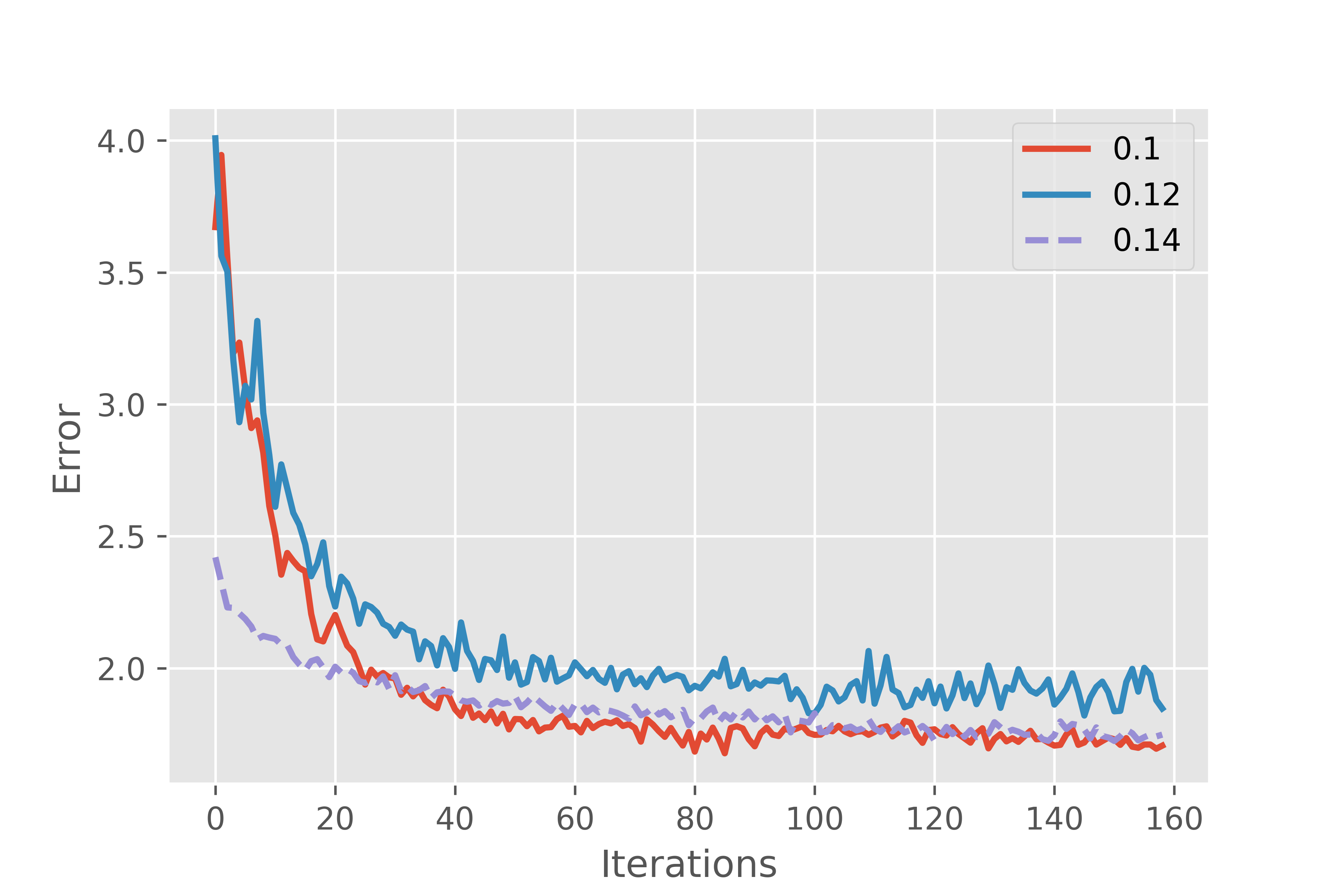}
            \caption[]
            {{\small CIFAR10 data sets  (10 Monte Carlo experiments)}}
        \end{subfigure}
        \caption[]
        {\small \textbf{The intervals of hermite coefficients}.}
        \label{fig:ActivationIntervals}
\end{figure}

The above analysis indicates that Hermite polynomials can be used to design useful activation functions, similarly with the use of wavelet for activation function designs \cite{Mallat2016}.
To get close to the edge of chaos, the Hermite coefficients of activation functions should be within $(0, 1)$.
In practice, to allow information to transfer in an iterative process, the coefficients should be around $0.5$ (the middle of the interval).
Also, based on the dynamics, the maximum and the minimum Hermite coefficients should also have an impact on the convergence. To demonstrate their impact, we plot the convergence error of DNNs on the MNIST and CIFAR10 data sets using different maximum and minimum Hermite coefficient values in
Figures \ref {fig:ActivationCompareStart} and \ref{fig:ActivationCompareEnd}.
The results show that for both data sets, when the maximum Hermite coefficient is \numrange{0.6}{0.65}, and the minimum is around $0.4$, the system is optimal.
In addition to the maximum and minimum Hermite coefficients, the spacing between the eigenvalues may also affect the performance of the Hermite polynomial (HP) activation function.
Fig. \ref{fig:ActivationIntervals} shows the influence of the interval of Hermite coefficients on convergence, indicating that the optimal interval is \numrange{0.12}{0.14}.

Through the above analysis, we find that, similarly with complex networks, there is an interesting spatiotemporal information representation and propagation in DNNs,
where information capability becomes more and more abundant by increasing spatial representation and the evolution towards the edge of chaos as depth grows \cite{hens2019spatiotemporal}.
The spatiotemporal structures can also be modeled by feature selection.
For example, instead of taking features only from current regions, elements from adjacent areas can be stacked together to form a larger feature space.
In the context of image processing, this idea was generalized to convolutional neural networks (CNNs) to calculate the adjacent properties of images \cite{KrizhevskySH17}.
RNNs provide another useful way to model the temporal dependency, rendering the network to extract the specific time-dependent length of information \cite{GravesMH13}.
%There is generally a tendency that spatiotemporal information representation for complex system modeling (see \cite{liu2017survey} for more details).

\section{Information transfer in deep neural networks: Training network models}\label{networksModelsTraining}
So far our discussion is mainly focused on the expressivity of DNNs. In this section, we discuss another crucial step, the training process. It is well known that different learning problems require different network models, and the training process plays a crucial-yet-challenging role \cite{LeeBNSPS18}. The training can be viewed as a complex matching process \cite{Mafahim2015Complexity}.
To date, many works involving model matching have focused on graph neural networks, since graphs can vividly capture the relationship between nodes \cite{HamiltonYL17}.
Xu et al. \cite{xu2018powerful} pointed out that the expressivity of GNNs was equivalent to the Weisfeiler-Lehman graph isomorphism test.
However, for general non-convex problems, solving the exact model matching is very difficult due to its NP complexity and a lack of guarantee for global optimality \cite{aflalo2015convex, bach2017breaking}.
Furthermore, current neural networks are invariably static networks, i.e., the network structures are pre-set before training.
Dauphin and other researchers have suggested that the number of saddle points (not local minimums) increases exponentially likely with the network depth \cite{dauphin2014identifying}.
In other words, a local minimum in a low dimension space may be a saddle point in a high dimension space. This phenomenon is also called by the proliferation of saddle points hypothesis, which can be easily verified by the dynamics.
From Sec. \ref{DNNBasics}, we know that DNNs with an activation function whose declining eigenvalues are within $(0, 1)$ at the output of the previous layer will result in unstable equilibrium points in the later layers.
Therefore exact model matching is rarely used in practical applications. Instead, an inexact matching is commonly used to relax the requirements with an error-tolerant measure.
For example, backpropagation is an inexact matching method by passing the information from a loss function to inputs based on calculating the derivative of the loss function.
Below we discuss more details on how the information transfers in backpropagation.

Backpropagation exploits a chain rule of information transfer using gradient-based optimization techniques.
The original gradient descent method provides an approximate linear convergence,
that is, the error between weights and loss function asymptotically converges in the form of a geometric series.
The convergence rate on the quadratic error surface is inversely proportional to the condition number of a Hessian matrix, i.e., the ratio between the maximum and the minimum eigenvalues $\lambda_{\max}/\lambda_{\min}$.
The corresponding eigenvectors represent different directions of curvature.
When the Hessian matrix has a high condition number, the eigenvalues attenuate at a high amplitude, and the corresponding derivative decreases rapidly in some directions, while declines gently in other directions.
Furthermore, when there is a saddle point (then the Hessian matrix is called an ill-conditioned matrix), it typically forms a long plateau with a gradient close to zero, which makes it difficult to escape from the area
(we have verified a similar point in the activation function analysis of Sec. \ref{HilbertFeatureMapping}).
This issue, however, becomes more complicated when there is a proliferation of saddle points in high dimensional DNNs.
Next, we discuss the latest development that may help alleviate this issue.

From the dynamics in   \eqref{eq:LinearizeAproximate}, the eigenvalues of gradient optimization are dependent on hidden weight matrices and inputs, besides the activation functions.
For arbitrary inputs, several methods such as principal component analysis can be adopted to ensure the input is convex (for graph data, the primary method is graph kernels \cite{lei2017deriving, oneto2017measuring}).
For convex networks, the first-order stochastic gradient descent method and its variants may break saddle points if the saddle points are along with other directions \cite{LiY17, gao2019conjugate}.
Some previous work has attempted to utilize second-order methods such as the Hessian matrix to adjust the direction obtained by first-order methods. But these methods only avoid local minimums, not saddle points \cite{dauphin2014identifying}.
Therefore, second-order methods are not always preferable to first-order methods.
Real-world practitioners often prefer a combinational algorithm that incorporates first-order methods and second-order methods, such as Adam.
In addition to the optimization methods, there is some effort dedicated to building the convex topological structures \cite{bu2019distributed}.
For example, Amos et al. \cite{amos2017input} developed a convex structure with a non-negative constraint on the hidden weight matrices.
Rodriguez et al. \cite{Rodriguez0CGR17}  proposed a regularization of CNNs through a decorrelated network structure.
Most recently, Zhang et al. \cite{ZhangSS19} confirmed via experiments that the success of recently proposed neural network architectures (e.g., ResNet, Inception, Xception, SqueezeNet, and Wide ResNet) was attributed to a multi-branch design that reduced the non-convex structure of DNNs.
However, most of these methods still focus on the local minimums; little is known for the saddle points that slow down the convergence.

In addition to optimization techniques, heuristic optimization techniques based on swarm intelligence are also explored to tackle the convergence of DNNs \cite{stanley2019designing}.
The most popular method was designed to pre-train the hyperparameters  (e.g., network depth, width, etc.) for later stage network setting \cite{Wang2015Back}.
Erskine et al. \cite{ErskineH15} proved the convergence of particle swarm optimization corresponding to self-organized criticality.
Moreover, Hoffmann et al. \cite{hoffmann2018optimization} demonstrated that criticality could be determined by the system's primary parameters.
One intuitive explanation is that if one can determine a moderate network model through prior training, then the saddle points caused by over-parameterization can be avoided.
Evolved topology also provides an exciting opportunity to address these challenges \cite{ZhangWCLMSB16}. However, they start with a small-scale network that takes a long time to achieve global optimality \cite{stanley2019designing}.
In conclusion, it is so far still unclear the ``best'' algorithm that can be employed to reliably analyze the convergence of DNNs, especially with a small computation budget.
Later we will discuss how an evolutionary algorithm can help DNN arrive at the edge of chaos through numerical examples.

\section{Experiment: input perturbation and the regularization}\label{InputPerturbationandRegularization}
\begin{figure}[htbp]
 \centering
 \includegraphics[width= 0.33\textwidth]{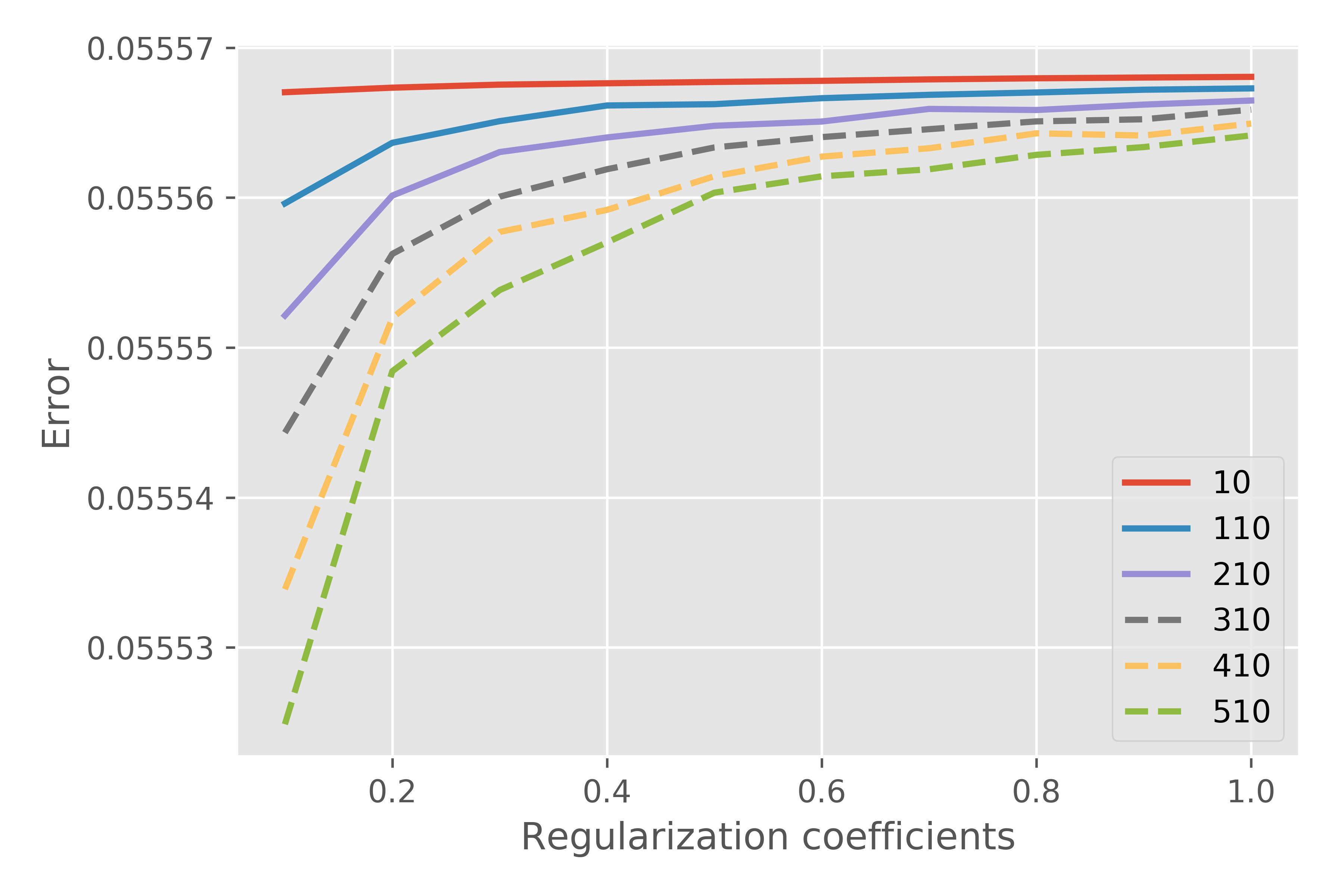}\\
 \caption{\textbf{Variation of $L_2$-regularization coefficients under different network sizes}.
The results are for multivariate time series prediction based on the Mackey-Glass time series. The ideal network size is in \numrange{400}{500}.
As can be seen, the error increases as the regularization coefficient increases.
This suggests that $L_2$-regularization is not completely necessary for performance improvement.}
\label{fig:RegularizationCompare}
\end{figure}

This section studies the impact of input perturbation and regularization operator on expressivity. According to the criticality theory, given a perturbation on the constant variance, e.g., Gaussian white noise variance,
DNN can eventually converge to a criticality via multilayer learning.
When there is an input abnormality after training, there will be a deviation to the critical point.
The offset, on the other hand, provides an indicator of whether an exception occurred or not \cite{climsysdzw004}.
%So is the variance-like regularizations, although they have proved to be very effective,
The offset created by the regularizations hinders the trend toward criticality and finally degrades the system's representation.
Fig. \ref{fig:RegularizationCompare} shows the variation of the $L_2$-regularization coefficients under different network sizes.
The result indicates that $L_2$-regularization may not be necessary since the error increases as the coefficients increase. Similar findings were also reported in  \cite{Rodriguez0CGR17}.

\section{Experiment: Expressivity and model training validity}\label{modelValidity}
This section examines the proposed expressivity and model training theory through the multivariate time series prediction and image classification examples.
First, we consider the evolution of time series prediction.

\subsection{The evolution of deep echo state networks for time series prediction}

\begin{figure}
        \begin{subfigure}[b]{0.33\textwidth}
            \centering
            \includegraphics[width=\textwidth]{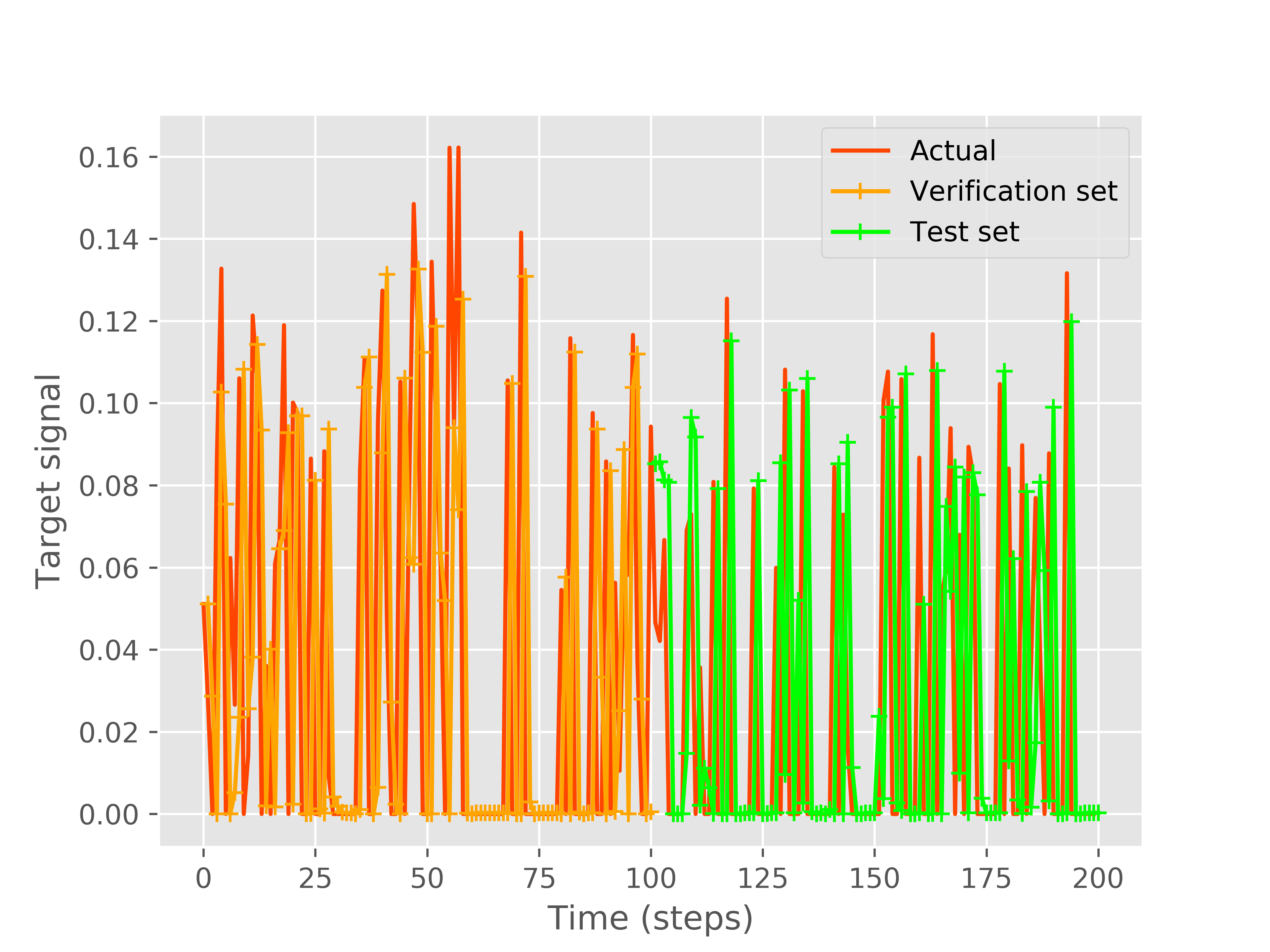}
            \caption[Network2]%
            {{\small Solar energy data set (0.7:0.1:0.2)}}
        \end{subfigure}
        \begin{subfigure}[b]{0.33\textwidth}
            \centering
            \includegraphics[width=\textwidth]{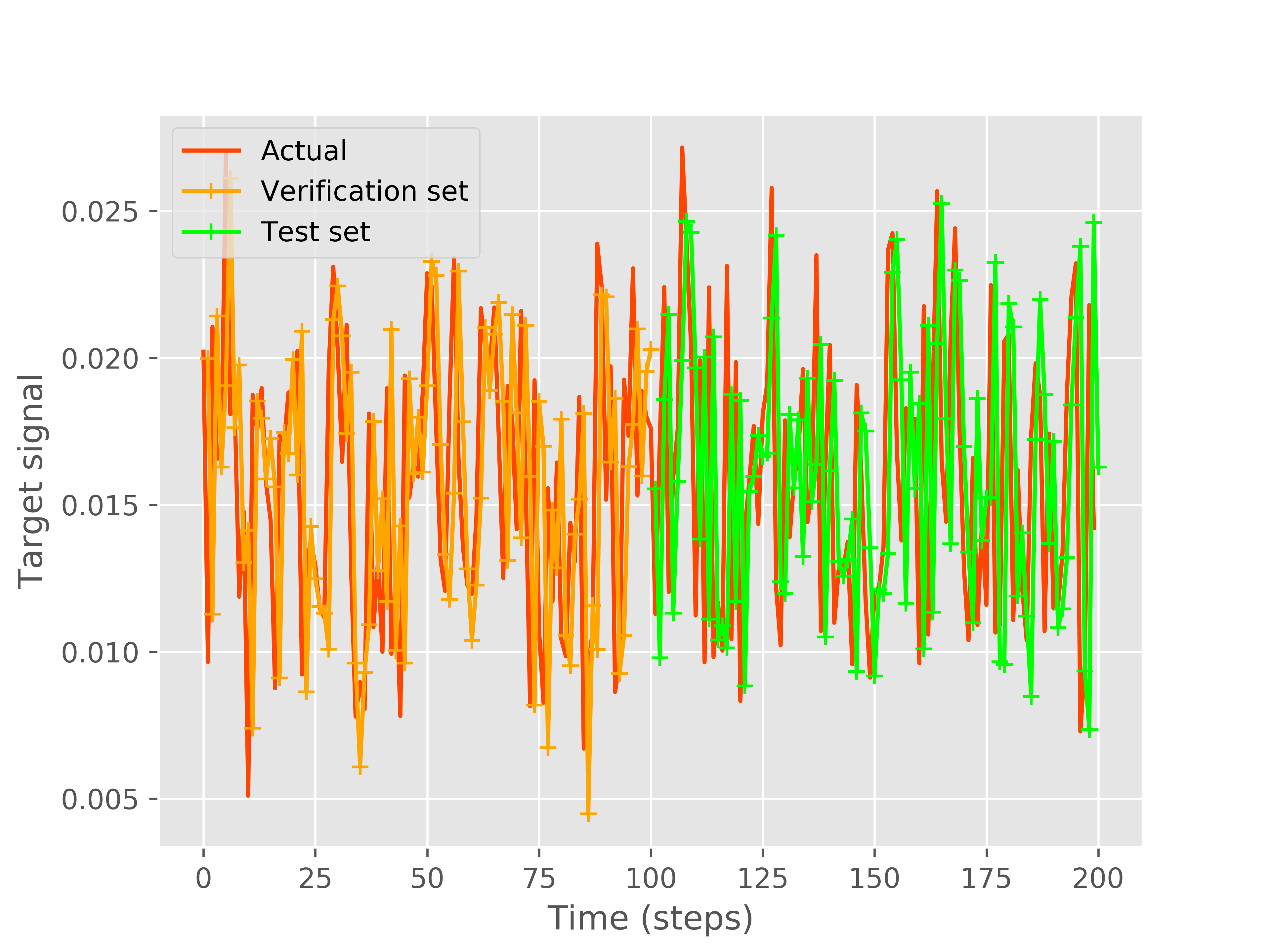}
            \caption[]%
            {{\small Traffic data set (0.7:0.1:0.2)}}
        \end{subfigure}
        \vskip\baselineskip
        \begin{subfigure}[b]{0.33\textwidth}
            \centering
            \includegraphics[width=\textwidth]{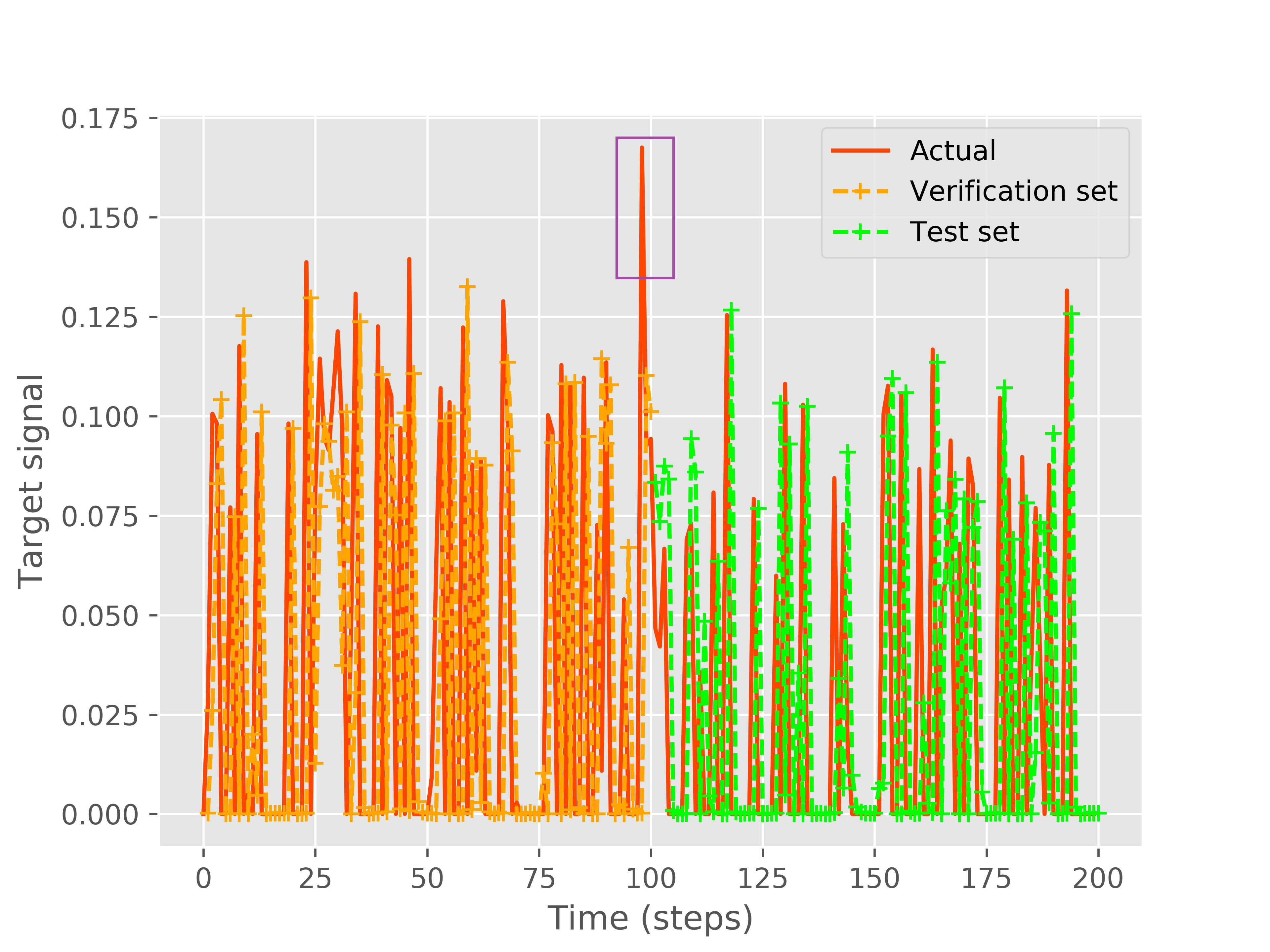}
            \caption[]%
            {{\small Solar energy data set (0.5:0.1:0`.4)}}
        \end{subfigure}
        \begin{subfigure}[b]{0.33\textwidth}
            \centering
            \includegraphics[width=\textwidth]{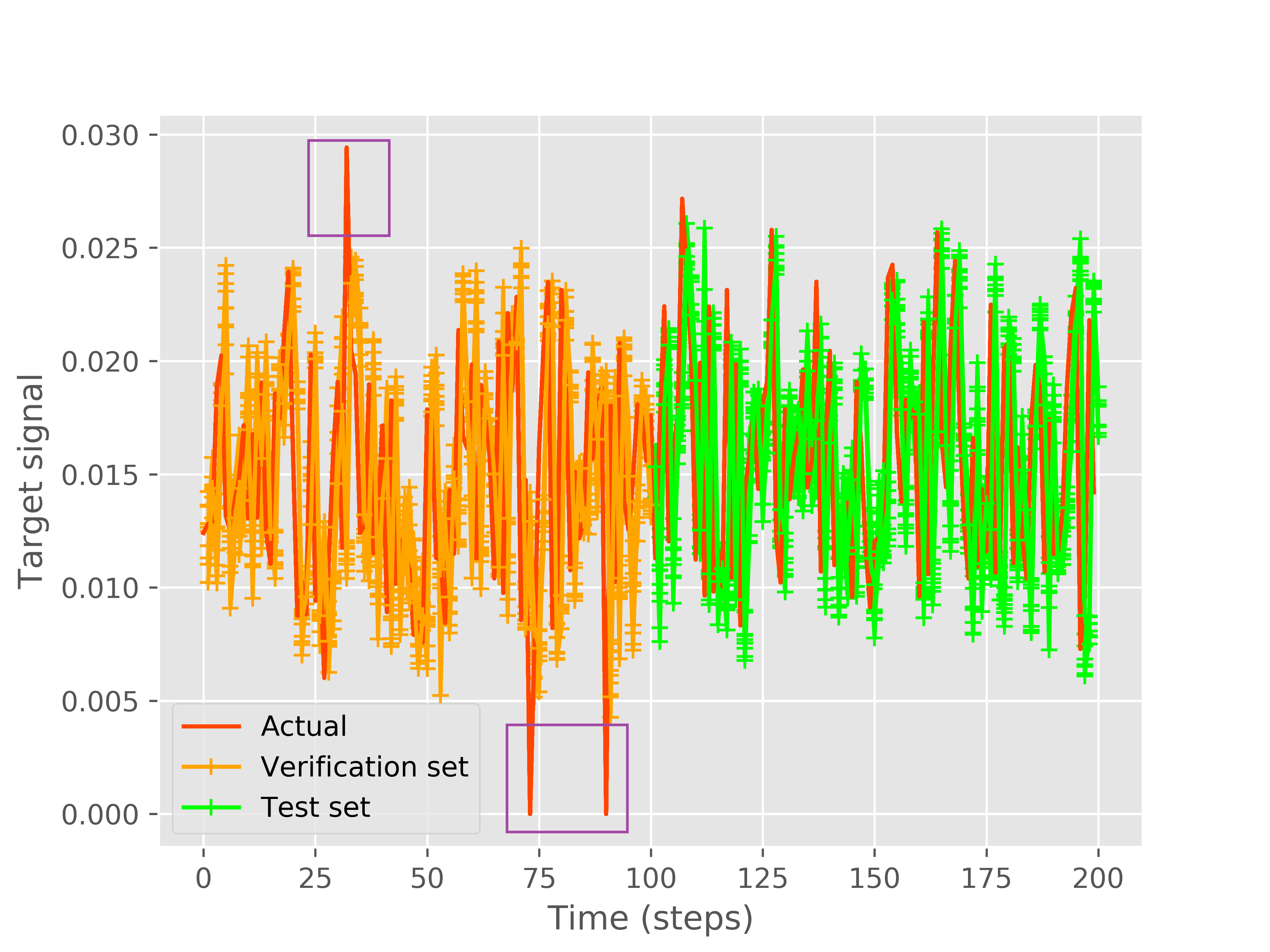}
            \caption[]%
            {{\small Traffic data set (0.5:0.1:0.4)}}
        \end{subfigure}
        \caption[]
        {\small Short-term prediction and long-term prediction resutls}
        \label{fig:PredictionResults}
\end{figure}
We consider two data sets for multivariate time series prediction purpose. The first data set is called Solar-Energy\footnote{\url{http://www.nrel.gov/grid/solar-power-data.html}}, where the solar power production in 2006 were obtained from 137 PV plants, and recorded every 10 minutes in Alabama, USA. The second data set is called
Traffic\footnote{\url{http://pems.dot.ca.gov}}, which is a collection of 48-month (2015-2016) hourly transportation data from California that describes the road occupancy rates measured by different sensors on San Francisco Bay Area freeways.

All data sets were chronologically divided into training sets, validation sets, and test sets.
The ratios of the training set and test set represent relatively short-term and long-term predictions, and the rate of validation set is fixed at $0.1$.

Fig. \ref{fig:PredictionResults} presents the multivariate time series prediction results via an improved DeepESN \cite{gallicchio2017deep},
in which activation is the HP activation, and the hyperparameters (i.e., network depth and width) are optimized by an evolutionary algorithm called population-based algorithm  (MPSOGSA) as proposed in \cite{zhang2018new}.
The optimized network layers for both data sets are $3$ and $6$, respectively.
The results show that the predictive accuracy of the test set is comparable to that of the validation set.
Besides, in subfigures (c) and (d), the proposed method is able to identify the potential anomalies/outliers (marked with purple boxes) since the prediction error exceeds a certain threshold.

\begin{figure}
        \centering
        \begin{subfigure}[b]{0.5\textwidth}
            \centering
            \includegraphics[width=\textwidth]{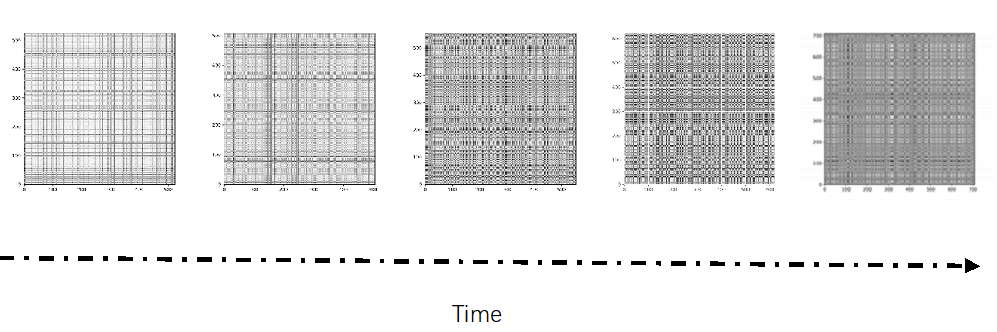}
            \caption[]i
            {{\small Solar energy data set}}
        \end{subfigure}
        \begin{subfigure}[b]{0.5\textwidth}
            \centering
            \includegraphics[width=\textwidth]{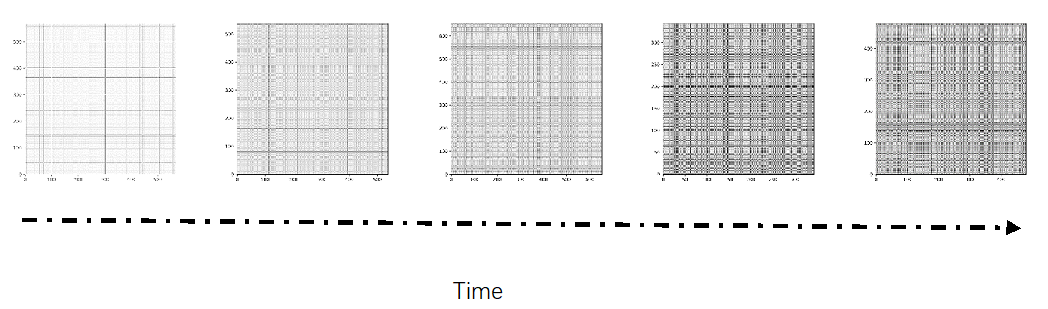}
            \caption[]
            {{\small Traffic data set}}
        \end{subfigure}
        \caption[]
        {\small \textbf{Evolution of the time dependence of hidden state}.}
        \label{fig:hiddenStateEvolution}
\end{figure}

Next, we examine the relationship between expressivity and training.
The hidden state of DNNs enables us to store information efficiently from the past, commonly referred to as the ``memory" of  DNNs \cite{yang2016new}.
One way to visualize this characteristic is the recurrence plot, which is a tool that displays recurrent states in phase space via a two dimensional plot \cite{Bianchi2018Investigating}. Consider
 \begin{equation}\label{eq:recurrencePlots}
{R_{i,j}} = \Theta ({\varepsilon _i} - |{\vec x_i} - {\vec x_j}|),\; {\vec x_i}, {\vec x_j} \in {\mathbb{R}^m}  \; i,j = 1, \dots, N,
\end{equation}
where $\Theta(\cdot)$ is the Heaviside step function, $N$ is the number of states, and $\varepsilon_i$ is the distance threshold.
When $|{\vec x_i} - {\vec x_j}| <{\varepsilon_i}$, the value of the recurrence matrix $R_{ij}$ is one; otherwise zero.
From an information transfer perspective, when the dynamics satisfy $ \bm{x}(n)=f(\bm{W}\bm{x}(n-1)) $, a strong time dependency appears and the system is on the edge of chaos.
Therefore, the closeness of the chaotic edge can be identified by the presence of a large number of black points in the recurrence plot.
Fig. \ref {fig:hiddenStateEvolution} shows the hidden state change from initial evolution to the final stage.
Indeed we see that as the hyperparameters of DeepESN evolve, considerable time dependencies appear\footnote{Sometimes there might be a convergence issue for the current evolutionary algorithm. As a result, we did not always observe the darkening effect over time in evolution, but the overall trend is quite obvious}.
Otherwise, if the network size and input do not match, there is no apparent time dependency.
This result verifies that suitable network size is essential for reaching the chaotic edge.
Based on the analysis in Sec. \ref{networksModelsTraining}, we can say that the training process is close to a chaotic edge, but whether it can ultimately arrive that edge depends on its ability to overcome convergence and to pass information to the required network depth.

\begin{table*}
    \begin{subtable}[h]{1\textwidth}
        \centering
        \begin{tabular}{l||lll|lll|lll}
\toprule \toprule
\multirow{2}{*}{Predictive methods} & \multicolumn{3}{|c|}{0.7:0.1:0.3} & \multicolumn{3}{c|}{0.6:0.1:0.2}&\multicolumn{
3}{c}{0.5:0.1:0.4}\\
\cline{2-10}
&\textbf{MAE} & \textbf{RMSE} & \textbf{MAPE} &\textbf{MAE} & \textbf{RMSE} & \textbf{MAPE}&\textbf{MAE} & \textbf{RMSE}
 & \textbf{MAPE} \\
\midrule
LSTNet \cite{LaiCYL18} &16.2569   &0.1338   &1.3332 &17.043   &0.1459   &1.4788 &19.9066&0.1659&1.2265\\
DeepESN \cite{gallicchio2017deep} & 16.9215   &0.1403   &1.4170& 17.1019&0.1586&1.5026&19.0810&0.1600    &1.0485 \\
HP-LSTNet &14.0910&0.1173 &1.2750 &15.4805 &0.1278 &1.2592 &17.7437   &0.1574&1.0128\\
HP-DeepESN &14.4248&0.1257 &1.3480 &15.8189&0.1264   &1.3560 &17.1191&0.1621   &1.1890 \\
Optimized-HP-DeepESN &13.4751 &0.1143 &1.1825 &14.1346 &0.1148 &1.2883 &15.7616 &0.1392 &0.8683\\
\hline
\end{tabular}
       \caption{Solar-Energy data set}
       \label{tab:week1}
    \end{subtable}
    \hfill
    \begin{subtable}[h]{1\textwidth}
        \centering
        \begin{tabular}{l||lll|lll|lll}
\toprule \toprule
\multirow{2}{*}{Predictive methods} & \multicolumn{3}{|c|}{0.7:0.1:0.2} & \multicolumn{3}{c|}{0.6:0.1:0.3}&\multicolumn{3}{c}{0.5:0.1:0.4}\\
\cline{2-10}
&\textbf{MAE} & \textbf{RMSE} & \textbf{MAPE} &\textbf{MAE} & \textbf{RMSE} & \textbf{MAPE}&\textbf{MAE} & \textbf{RMSE} & \textbf{MAPE} \\
\midrule
LSTNet \cite{LaiCYL18} &26.2603 &0.0638 &0.4561 &26.7099 &0.0645 &0.4498&27.6093 &0.0677 &0.4856 \\
DeepESN \cite{gallicchio2017deep} &26.0551&0.0641&0.4519&27.1709& 0.0695&0.5212&27.6028 &0.0657&0.4772\\
HP-LSTNet &25.6275 &0.0597 &0.4143 &25.804 &0.0610 &0.4846&27.1432 &0.0613 &0.4646 \\
HP-DeepESN &25.6869 &0.0617 &0.4202 & 25.6190 &0.0654 &0.4886&27.1368 &0.0607 &0.4613\\
Optimized-HP-DeepESN &24.7516 &0.0442 &0.3043&24.0963 &0.04252 &0.4063 &25.04305 &0.0424 &0.2588\\
\hline
\end{tabular}
        \caption{Traffic data set}
        \label{tab:week2}
     \end{subtable}
     \centering
     \caption{Performance comparison of different multivariate time series prediction methods.}
     \label{Tab:AlgoComp}
\end{table*}

The above analysis also shows that the time depends on the training process, which is closely relevant to the long-term time series prediction\footnote{A common issue in the long-term prediction is that time dependencies may be extremely long, for more details, refer to \cite {LaiCYL18}.}.
Next, we compare the performance of the given method with that of two benchmark methods (LSTNet \cite{LaiCYL18} and
DeepESN \cite{gallicchio2017deep}) to check if they can improve predictive performance, especially for long-term series prediction.
Tab. \ref{Tab:AlgoComp} shows a comparison result of different multivariate time series prediction in terms of mean absolute error (MAE), root mean squared error (RMSE), and mean absolute predictive error (MAPE) for different training/testing/validation proportions.
We observe that LSTNet achieves better results than DeepESN for short-term prediction of $(0.7:0.1:0.2)$,
while DeepESN is slightly better than LSTNet for long-term prediction of $(0.5:0.1:0.4)$.
Besides, the two baseline algorithms have a considerable improvement when the HP activation is utilized to capture spatial features.
The optimized HP-DeepESN, in general, has the best performance as it can overcome convergence issues.

\subsection{The impact of batch size on the convergence for image classification}
\begin{figure}
        \centering
        \begin{subfigure}[b]{0.475\textwidth}
            \centering
            \includegraphics[width=\textwidth]{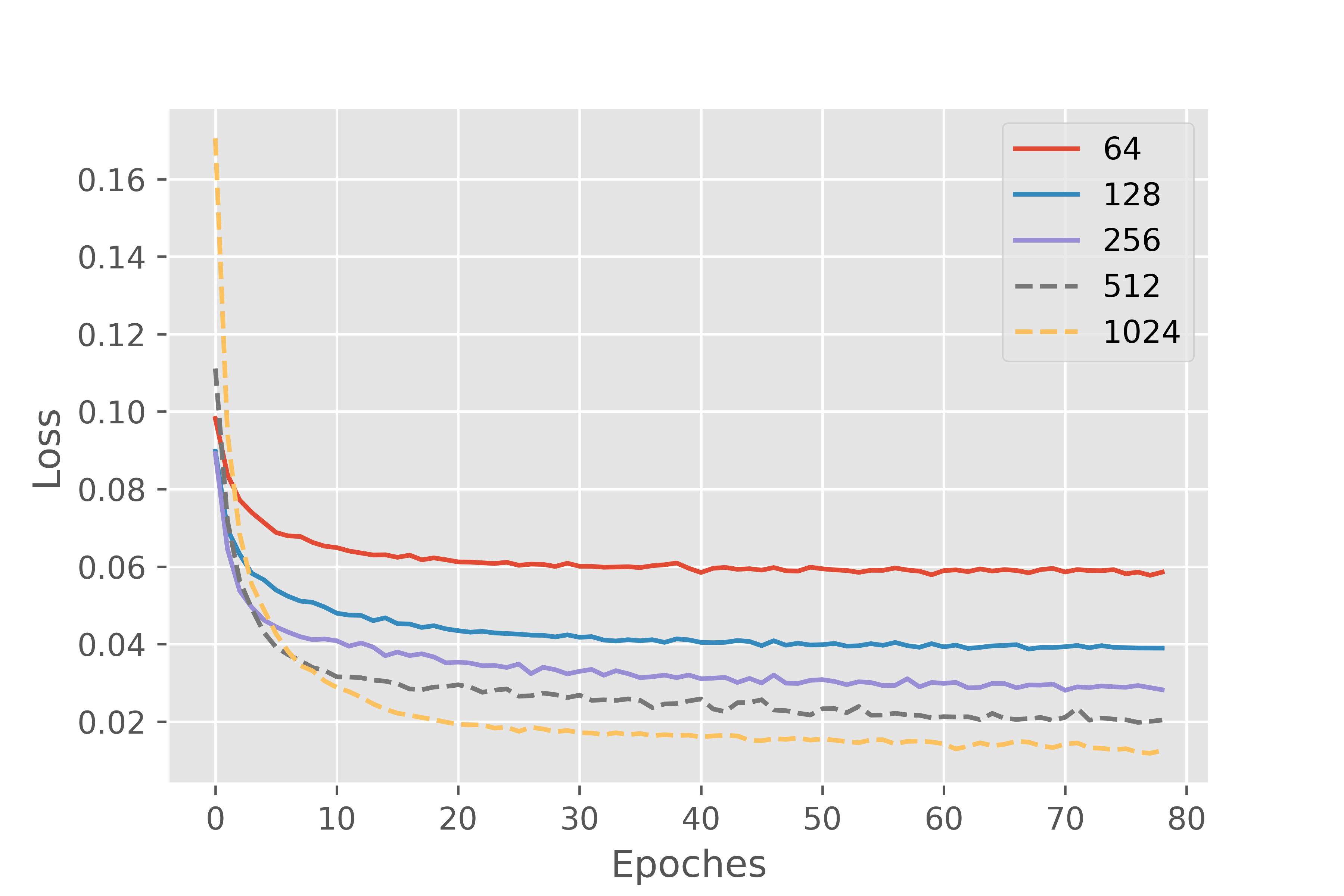}
            \caption[]
            {{\small MNIST data set  (10 Monte Carlo experiments)}}
        \end{subfigure}
        \begin{subfigure}[b]{0.475\textwidth}
            \centering
            \includegraphics[width=\textwidth]{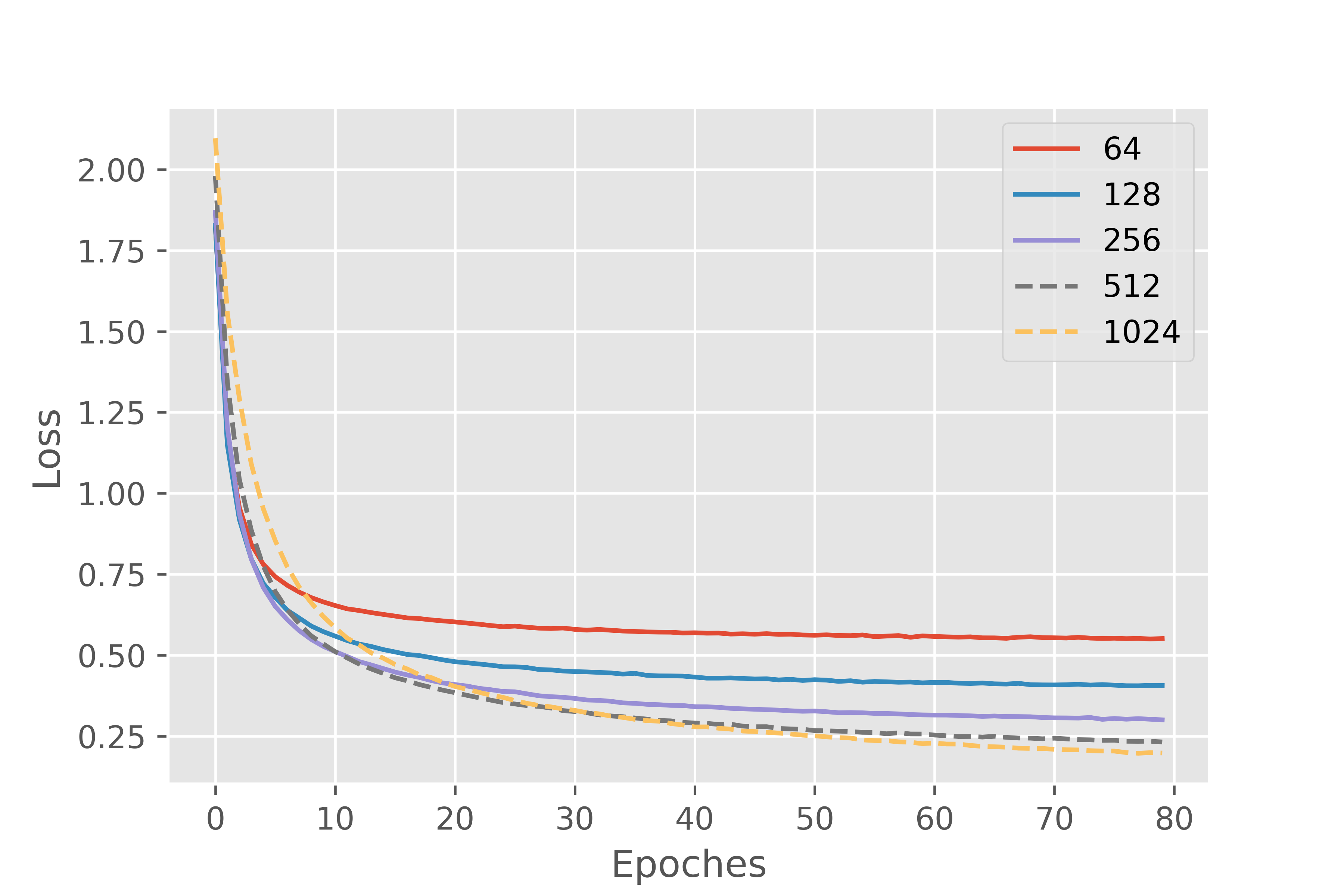}
            \caption[]
            {{\small CIFAR10 data set  (20 Monte Carlo experiments)}}
        \end{subfigure}
        \caption[]
        {\small \textbf{The impacts of batch size on the convergence of image classification}.}
        \label{fig:BatchSizeChangesConvergence}
\end{figure}

\begin{figure}
        \centering
        \begin{subfigure}[b]{0.5\textwidth}
            \centering
            \includegraphics[width=\textwidth]{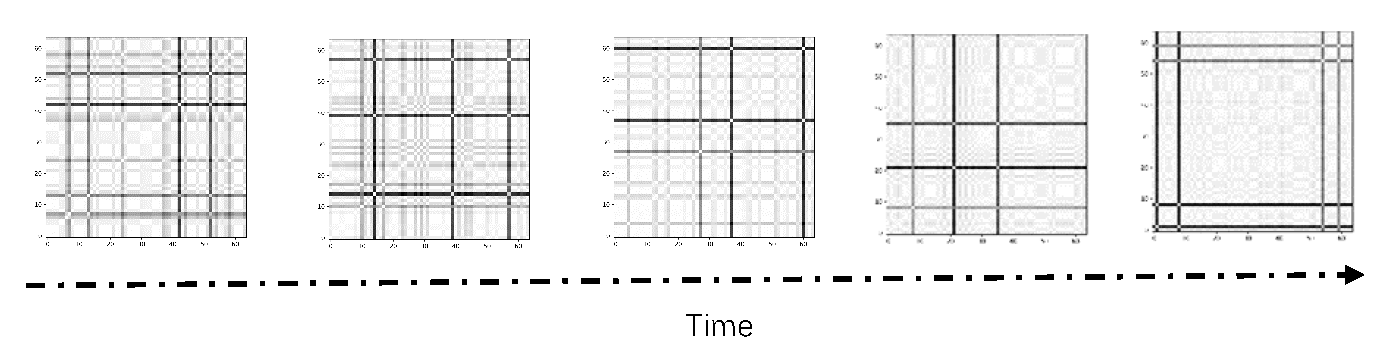}
            \caption[]
            {{\small MNIST data set}}
        \end{subfigure}
        \begin{subfigure}[b]{0.5\textwidth}
            \centering
            \includegraphics[width=\textwidth]{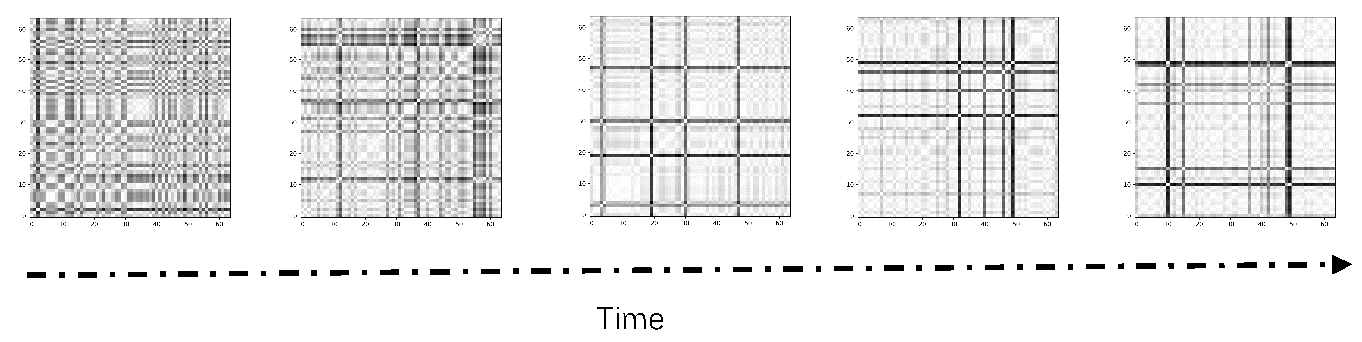}
            \caption[]
            {{\small CIFAR10 data set}}
        \end{subfigure}
        \caption[]
        {\small \textbf{Evolution of the time dependence of hidden state}.}
        \label{fig:ImageHiddenStateEvolution}
\end{figure}

Next, we analyze the impact of batch size on information transfer for image classification.
Fig. \ref{fig:BatchSizeChangesConvergence} shows the effect of batch size on convergence based on the MNIST and CIFAR10 data sets.
 As the batch size increases, the loss converges. Meanwhile, as shown in  Fig. \ref{fig:ImageHiddenStateEvolution}, the hidden state is gradually disappearing,
 which means the information transfer becomes weak. Therefore this example gives an illustration that an appropriate batch size should be selected to balance the trade-off between the convergence of loss and the information transfer strength.

\section{Related work}\label{relatedWorks}
In the literature, the mean-field theory has been widely used to describe the expressivity of neural networks \cite{poole2016exponential, ChenPS18, SchoenhiolzGGS17, XiaoBSSP18, YangPRSS19}. For example, several groundbreaking studies using statistical physics have shown that DNNs exhibit exponential expressivity as network depth increases.
These works highlighted the importance of orthogonal weights initialization \cite{poole2016exponential, SchoenhiolzGGS17, XiaoBSSP18}, and showed that the gating mechanism could help RNNs avoid local minimums \cite{ChenPS18}.
Besides, Schoenhiolz and Yang et al. have found that dropout or batch normalization is also a cause of gradient explosion \cite{SchoenhiolzGGS17, YangPRSS19}.
Compared to those previous work based on mean-field theory, our work manages to obtain similar results in terms of expressivity without the need of certain assumptions in statistical physics theory, Moreover, our work highlights the connection between the actual training and expressivity, which has not been covered by field theory.

\section{Concluding remarks}\label{Conclusions}
This paper discusses the expressivity and training issues of DNN  based on its convergence and criticality.
A dynamic model in Hilbert space is employed to analyze the feature mapping of a vanilla DNN.
By studying the feature mapping of several significant activation functions, we find that the eigenvalues in the feature space have sharp decay or even saddle points, hence slowing down the information transfer in DNNs.
An activation function design based on Hermite polynomials is then proposed to make better use of the spatial representation.
Training issues, especially the convergence problem caused by dimensional mismatching between inputs and network models, involving backpropagation are analyzed.
The impact of input perturbation and regularization operators on expressivity is also investigated.
The analysis shows that in theory, DNN uses spatial domains for information representation and evolves to the edge of chaos as depth increases, where (shannon) entropy reach a maximum.
In actual training, whether a network can arrive or not depends on its ability to overcome convergence and pass information to the required network depth.

Model matching between input and output topological structure is critical to facilitate fast convergence.
For DNNs, this is achieved by adding or removing neuron connections.
Through the formation of saddle points, the network structure needs to have desired contraction characteristics to avoid saddle points in high dimensional representation \cite{Mallat2016}.
Therefore, it will be of interest to explore the efficient dynamic topology evolution in future work to help provide guidelines for designing networks with faster convergence.

\appendix

\section{Hermite polynomials}\label{ActivationinHermitePolynomials}
This part includes the definition of the Hermite polynomial and the properties we used.
Supposing the probabilists's weight function $p(x) = e^{-x^2/2}$, apply Lemma 1 it follows that the Hermite polynomials are orthogonal with respect to the weight function in the interval $(-\infty, \infty)$,
 then we have the following important results,
 \begin{equation}\label{eq:HermtiePolynomial}
 \small{
 \int_{-\infty}^{\infty} H_m(x) H_n(x) e^{-x^2/2}\text{d}x=
 \begin{cases}
 2^n n!\sqrt{\pi},& \text{ for } m = n\\
 0      , &\text{otherwise}.
 \end{cases}}
 \end{equation}

We use the following facts about the Hermite polynomials (see Chapter 11 in \cite{o2014analysis}):
\begin{align}
H_{n+1}(x)&= \frac{x}{\sqrt{n+1}}H_n(x)- \sqrt{\frac{n}{n+1}}H_{n-1}(x),\\
H_n^{\prime}(x)&=\sqrt{n}H_{n-1}(x),
\end{align}
\begin{equation}
H_n(0)= \begin{cases}
0, & \text{if } n \text{ is odd};\\
 \frac{1}{\sqrt{n!}}(-1)^{\frac{n}{2}}(n-1)!! & \text{if } n \text{ is even}.
\end{cases}
\end{equation}

Next, we analyze the eigenvalues of several commonly used activation functions under Hermite polynomials and their impacts on the convergence and criticality of DNNs.
\subsubsection{Radial basis function activation}
If the activation used is RBF, the corresponding feature space is a Hilbert space of infinite dimension, whose value depends on the distance from the fixed point $c$.
\begin{equation}\label{eq:expFunc}
\small{
\mathop{f}(x) = \exp(-\frac{\lVert x-c\rVert^2}{2\sigma^2})}.
\end{equation}
where $\sigma$ is a scale parameter.
Since $H_0(x) = 1, H_1(x) = x, \text{ and } H_2(x) = \frac{x^2-1}{\sqrt{2}}$,
Substitute Eq. (\ref{eq:expFunc}) into Eq. (\ref{eq:HermtiePolynomial}) , and get the corresponding Hermite coefficients:
\begin{align}
 a_0&= \frac{\sqrt{2\pi}\sigma c e^{- c^2/(2 \sigma^2 +2)}}{\sqrt{\sigma^2 + 1}},\\
 a_1&= \frac{\sqrt{2 \pi} c e^{- c^2/(2 \sigma^2 +2)}}{\sigma(\sigma^2 + 1)^{3/2}},\\
 a_2&= \frac{\sqrt{2 \pi} c e^{- c^2/(2 \sigma^2 +2)}}{\sigma(\sigma^2 + 1)^{5/2}}.
\end{align}
Therefore, the Hermite coefficients of RBF can be expressed as:
\begin{equation}
a_n= \frac {\sqrt {2 \pi} ce ^ { - c^2/(2 \sigma^2 +2)}} {\sigma(\sigma^2 + 1)^ {(n+1/2)}}, \; n=0, 1, \dots.
\end{equation}
We see all eigenvalues $ a_n> 0, n = 0, 1, \dots$, indicating that RBF activation limits quadratic error surfaces that converges to the global minimum (or maximum).
Besides, the coefficients attenuates at the speed of $ \frac{a_n} {a_{n-1}} = \frac {1} {\sigma^2 + 1} $.
If $\sigma^2\leq 1$, the coefficients attenuate slowly; otherwise, the coefficients decay rapidly.
It is, therefore, that small $\sigma$ should be chosen as activation, and Shi et al. have made the RBF as activation function \cite{ShiFH18}.

\subsection{Step activation}
According to the literature \cite{DanielyFS16}, the Hermite polynomials of Step activation are:
\begin{equation}
a_n = \begin{cases}
\frac{(n-2)!!}{\sqrt{2\pi n!}} & \text{if } n \text{ is odd};\\
\frac{1}{\sqrt{2}} & \text{if } n= 0;\\
 0 &\text{if } n \text{ is odd} \geq 2.
   \end{cases}
\end{equation}

\subsection{ReLU activation}
The Hermite coefficients of ReLU activation are \cite{DanielyFS16}:
\begin{equation}
a_n = \begin{cases}
\frac{(n-3)!!}{\sqrt{\pi n!}} & \text{if } n \text{ is even};\\
\frac{1}{\sqrt{2}} & \text{if } n= 1;\\
 0 &\text{if } n \text{ is odd} \geq 3.
   \end{cases}
\end{equation}

The maximum eigenvalue is $\frac{1}{\sqrt{2}}$, then gradually decay to the critical point $0$.
We also see that there are $0$ values with an interval of $1$,
which may result in information that can not be efficiently passed from inputs to outputs, nor is backpropagation, it is unfavorable from the perspective of information transfer \cite{LiY17}.

\subsection{Sigmoid activation}
Consider the Sigmoid function $ \sigma(x)= \frac {1} {e ^ {-x} + 1}$, substitute it into Eq. (\ref{eq:HermtiePolynomial}), and get:
\begin{align}
   a_0 & =\mathbb{E}_{x\sim N(0,1)} [\sigma(x)]=\frac{1}{2},\\
   a_1 &= \mathbb{E}_{x\sim N(0,1)} [\sigma(x)x]=0.206621, \\
   a_2 & = \frac{1}{\sqrt{2}}\mathbb{E}_{x\sim N(1,1)} [\sigma(x)(x^2-1)]=0
\end{align}
According to the symmetry of integral operator, when $n$ is even, the Hermite polynomial is $0$.
We know that Sigmoid activation attenuates the values with a higher magnitude.
Since the first Hermite coefficients of sigmoid activation is much smaller than $1$ and saturate quickly, when it is used as an activation function in DNNs, the chance of the gradient fading to $0$ is very high.
%   \addtolength{\textheight}{-12cm}
\subsection{Swish activation function}
 Consider the activation $f(x)=x \text{sigmoid}(x)$. Substitute it into Eq. (\ref{eq:HermtiePolynomial}), and get:
\begin{align}
   a_0 & =\mathbb{E}_{x\sim N(0,1)} [\sigma(x)]= 0.292206\\
   a_1 &= \mathbb{E}_{x\sim N(0,1)} [\sigma(x)x]=\frac{1}{\sqrt{2}}\\
   a_2 & = \frac{1}{\sqrt{2}}\mathbb{E}_{x\sim N(0,1)} [\sigma(x)(x^2-1)]=0.350845
\end{align}

Calculation from Mathematical \cite{Mathematica}, we conclude that there is no $0$ value in the Swish activation, forming a convex function, and the coefficients decay slowly, so it can effectively pass information in DNNs, which can be used as an activation function.

\bibliographystyle{unsrt}
\bibliography{ReservoirChaosControl}

\end{document}